\documentclass[lettersize,journal]{IEEEtran}
\usepackage{amsmath,amsfonts}
\usepackage{algorithmic}
\usepackage{algorithm}
\usepackage{array}
\usepackage[caption=false,font=normalsize,labelfont=sf,textfont=sf]{subfig}
\usepackage{textcomp}
\usepackage{stfloats}
\usepackage{url}
\usepackage{verbatim}
\usepackage{graphicx}
\usepackage{cite}
\usepackage{multirow}
\newboolean{debug}
\setboolean{debug}{true} 
\usepackage{color}
\usepackage{cleveref}
\usepackage{tabularx, booktabs, pifont, threeparttable}
\usepackage{ulem}

\newcommand{\msout}[1]{%
    \ifthenelse{\boolean{debug}}{%
        \textcolor{red}{\sout{#1}}%
    }{%
    }%
}

\newcommand{\lia}[1]{%
    \ifthenelse{\boolean{debug}}{%
        \textcolor{red}{#1}
    }{%
        #1%
    }%
}

\newcommand{\lic}[1]{%
    \ifthenelse{\boolean{debug}}{%
        \textcolor{red}{#1}
    }{%
        #1%
    }%
}

\newcommand{\lib}[1]{%
    \ifthenelse{\boolean{debug}}{%
        \textcolor{yellow}{#1}
    }{%
        #1%
    }%
}
\newcommand{\zou}[1]{%
    \ifthenelse{\boolean{debug}}{%
        \textcolor{blue}{#1}
    }{%
        #1%
    }%
}

\newcommand{\li}[1]{%
    \ifthenelse{\boolean{debug}}{%
        \textcolor{red}{#1}
    }{%
        #1%
    }%
}

\begin{document}

\title{Simple but Stable, Fast and Safe: Achieve End-to-end Control by High-Fidelity Differentiable Simulation}

\author{Fanxing Li*, Shengyang Wang*, Yuxiang Huang*, Fangyu Sun, Shuyu Wu, Yufei Yan, Danping Zou, Wenxian Yu
\thanks{Fanxing Li, Yuxiang Huang, Fangyu Sun, Yufei Yan are with Shanghai Jiao Tong University({li.fanxing, crosshill, sunfly\_cc, sawyer\_wu, yanyufei, dpzou, wxyu}\@sjtu.edu.cn). Shengyang Wang(sw592\@duke.edu) is with Kunshan Duke University.}
}
\markboth{Journal of \LaTeX\ Class Files,~Vol.~14, No.~8, August~2021}%
{Shell \MakeLowercase{\textit{et al.}}: A Sample Article Using IEEEtran.cls for IEEE Journals}
\maketitle

\begin{abstract}

Obstacle avoidance is a fundamental vision-based task essential for enabling quadrotors to perform advanced applications. When planning the trajectory, existing approaches both on optimization and learning typically regard quadrotor as a point-mass model, giving path or velocity commands then tracking the commands by outer-loop controller.
However, at high speeds, planned trajectories sometimes become dynamically infeasible in actual flight, which beyond the capacity of controller.
Although direct taking low-level bodyrate commands as output can mitigate this issue, it gets much challenging to design such a low-level policy because the transition process is so complex and less smooth that the difficulty of training significantly increases.
In this paper, we propose a novel end-to-end policy that directly maps depth images to low-level bodyrate commands by reinforcement learning via differentiable simulation.
The high-fidelity simulation in training after parameter identification significantly reduces all the gaps between training, simulation and real world.
Analytical process by differentiable simulation provides accurate gradient to ensure efficiently training the low-level policy without expert guidance. 
The policy employs a lightweight and the most simple inference pipeline that runs without explicit mapping, backbone networks, primitives, recurrent structures, or backend controllers, nor curriculum or privileged guidance. By inferring low-level command directly to the hardware controller, the method enables full flight envelope control and avoids the dynamic-infeasible issue.
Experimental results demonstrate that the proposed approach achieves the highest success rate and the lowest jerk among state-of-the-art baselines across multiple benchmarks. The policy also exhibits strong generalization, successfully deploying zero-shot in unseen, outdoor environments while reaching speeds of up to 7.5m/s as well as stably flying in the super-dense forest.
Furthermore, we provide the first successful demonstration of backpropagating image information through high-fidelity differentiable simulation, validating the extensibility of first-order gradient methods to other complex robotic systems.
\end{abstract}

\begin{IEEEkeywords}
Collision-free flight, Backpropagation-through-time, High-field differentiable simulation, End-to-End.
\end{IEEEkeywords}

\section{Introduction} 
\label{sec:introduction}

Quadrotors have successfully perform various of high-risk application in real world like rescuing, searching, and delivering. Collision-free is the fundamental function to ensure safety so that it could execute other high level tasks. 
Conventionally, a typical technical pipeline of obstacle avoidance is comprised of three components: perception\&mapping, planning and control. Multiple procedures lead to accumulated error as well as excessive hardware overload \cite{hanover_autonomous_2024, zhou_ego-planner_2021, b_zhou_robust_2019}. Consequently, in real world with limited onboard computing resources, they usually fail in high-speed flight due to computation delay.

Therefore, researchers employs neural networks to replace one or several time-consuming modules, which could significantly accelerate the inference speed by omitting complex calculations like mapping and planning. Quadrotors could imitate expert planners with privileged knowledge \cite{jung_perception_2018, kaufmann_beauty_2019, cabrera-ponce_gate_2019} or learn from trial-and-error in reinforcement learning (RL) framework \cite{yu_mavrl_2024}. By reducing the inference delay, learning-based methods could achieve much faster speed than conventional methods \cite{loquercio_learning_2021} in outdoor environments. Exceptionally, there are also plenty of works that incorporate learning techniques like LSTM \cite{yu_mavrl_2024, bhattacharya_vision_2025} or backbone \cite{lu_yopov2-tracker_2025, loquercio_learning_2021} to enhance the inference capacity, which enable quadrotors to navigate in varying-density obstacle environments \cite{yu_mavrl_2024} as well as mazes \cite{han_hierarchically_2025} rather than cluttered forests.

However, although learning-based methods could accelerate inference to give stable-frequency commands, it becomes challenging and still probably fails in high-speed flight because of dynamics-infeasibility. 
Specifically, most of current works on collision-free flight, \cite{zhang_back_2024, loquercio_learning_2021}, use trajectory or velocity representatives as the output of actor network, with a following controller to transit it into low-level collective thrust and bodyrates (CTBR) command and send to hardware controller PX4 or BetaFlight. In training environment, such pipeline treats quadrotors as a point-mass kinematic model, which inherently leads to the training-to-inferring gap. When deployed in the much complex high-fidelity 6 degree of freedom (DOF) environment, if the attitude of quadrotors gets much extreme at high-speed, it becomes difficult to track desired trajectories within tolerated error due to dynamic restraints. 
Although such dynamics-infeasibility could be mitigated by finetuning the trajectory using optimization techniques like time-reallocation \cite{b_zhou_robust_2019, richter_polynomial_2016}. But if introducing another component, this excessive computation increases the inference duration, conflicting with the initial time-saving aim raised by learning-based methods. 

An alternative resolution is directly inferring the CTBR command with much better control response than position or velocity command. However, training such policies has two difficulties. On the one hand, if taking CTBR as command, it becomes challenging to estimate the long-time return in RL training,  because the transition process gets much complicated by introducing the high-fidelity dynamics, thus it has to rely on external guidance\cite{y_song_learning_2023} like imitation learning. On the other hand, movement direction control couples with orientation, improving the difficulty to constantly keep perception aware as well as follow desired movement, or it probably crashes due to losing view of approaching obstacle.

Recently, RL via differentiable physics emerges and exhibits improved convergence properties in quadrotor tasks \cite{li_abpt_2025, heeg_learning_2024}, as it could obtain analytical gradient through accumulative rewards from differentiable physics rather than using the critic to approximate it. It firstly achieves no-map high-speed flight in outdoor environment without expert guidance. Although currently research involving differentiable physics focus on point-mass model, the inherent differentiability of dynamics still shows promising extensibility that could be transferred onto high-fidelity robot system.

Therefore, in this paper, we present the first end-to-end policy for high-speed collision-free flight with low-level control commands. The low-level control avoids the dynamics-infeasibility by directly outputs CTBR. We leverage the high-fidelity simulation aligned with real-world robots, to train the policy with Back-Propagation-Through-Time (BPTT) via differentiable high-fidelity simulation, tackling the training efficiency training issue by introduced complex dynamics as well as adapting the altitude to achieve perception-aware. The analytical gradient provided by differentiable simulation enhances the learning efficiency, which makes even a 6-layer network capable to learn the policy, omitting the complex network architecture like LSTM and backbones. Meanwhile, direct inferring low-level command makes inference pipeline git rid of all the supportive controllers, makes the policy the simplest but efficient among current collision-avoidance research.  To the best of our knowledge, it is the first example that backpropagates the image features upon the gradient graph of high-fidelity \textbf{dynamics} rather than simplified \textbf{kinematics} in previous works.

On the premise that trained without any advanced architecture including recurrent network, backbone, action primitive, or external controller, or training tricks like privileged guidance or curriculum,  our policy exhibits the smoothest trajectory as well as the highest success rate among all baselines in a series of simulated experiments with increasing obstacle density and desired velocity. We have zero-shot deployed the policy in both urban and wild unseen real-world scenarios, which shows strong generalization as well as safety. It reaches the speed of 7.5 m/s in dense forest, same as the state-of-the-art algorithms. Within sequential aggressive maneuvers and varying environment, it could automatically adjust the velocity according to dynamic limitation to ensure safety when flying in super-dense forest. 

Our contribution could be summarized as:

\begin{itemize}
    \item We propose the first end-to-end policy that directly outputs CTBR commands in high-speed obstacle avoidance task, theoretically tackling the dynamic-infeasible problem that the tracking controller fails to track given trajectory from policy in extreme cases.
    
    \item We provide the simplest training and inferring pipelines in learning methods, which performs without any recurrent architecture, decoder-encoder, LSTM, backbone, action primitives or additional controllers.

    \item We conduct a series of experiments in simulation and real world. The results show that, even with the simple pipeline, our policy still exhibits the highest success rate as well as the smoothest path.
    
    \item We prove that, the image features are able to be backpropagated in limited horizon comprised of high-fidelity differentiable dynamics, even with much longer and complex gradient graph.
\end{itemize}

\begin{figure*}
    \centering
    \includegraphics[width=1.0\textwidth]{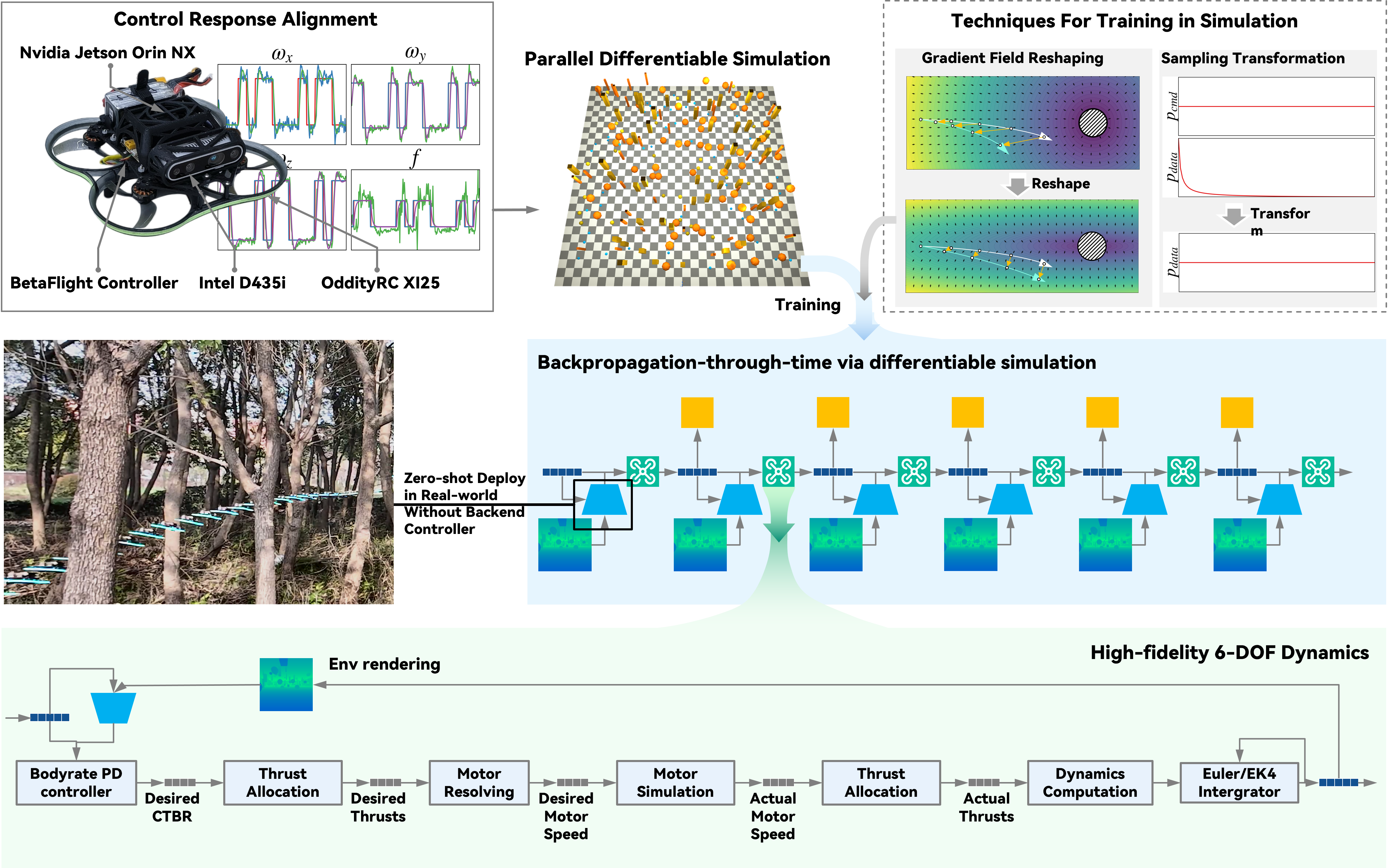}
    \caption{This section provides an overview of the proposed method. The policy is trained using backpropagation-through-time (BPTT) within a differentiable simulation that is aligned with real-world control responses. The high-fidelity simulation incorporates a series of precise dynamic computations, enabling successful policy training through straightforward training and inference pipelines, as well as zero-shot deployment on real-world robots.}
    \label{fig:overview}
\end{figure*}

\section{Related Works}
\subsection{Collision-free Flight}
Safe flight has always been a primary concern in aerial robotics. Early research focused on developing robust and efficient control techniques for precise trajectory tracking. As control research matured and converged, collision-free flight was first decoupled into mapping, planning, and control modules, and this decomposition has since served as the standard classic workflow.
In this pipeline, depth images or point clouds are converted into voxel grids or occupancy maps, collision-free trajectories are then planned based on the map, and finally, the planned trajectory is tracked using a robust controller. Planning methods generally fall into three categories: search-based methods \cite{s_liu_search-based_2018}, sampling-based methods \cite{allen_real-time_2016, richter_polynomial_2016}, and optimization-based methods \cite{b_zhou_robust_2019, zhou_ego-planner_2021}. Among these, optimization-based methods have been widely adopted in recent years due to their efficiency and smoothness. They have successfully addressed the challenge of real-time planning in low-speed scenarios for both single- and multi-robot systems \cite{zhou_ego-swarm_2021}. However, these methods still suffer from odometry drift and mapping uncertainty in high-speed scenarios, which can lead to inevitable collisions. Furthermore, the multi-module pipeline introduces additional complexity and computational cost, making it difficult to deploy onboard small-scale aerial robots with limited computational resources.

To reduce hardware overload and errors caused by traditional pipelines, researchers have employed neural networks to replace one or several components within the pipeline using deep learning \cite{jung_perception_2018, kaufmann_beauty_2019, cabrera-ponce_gate_2019, kouris_learning_2018}. The first category of policies involves mimicking expert actions generated by manual guidance annotations \cite{gandhi_learning_2017, giusti_machine_2015, loquercio_dronet_2018}, where professional algorithms have access to privileged observations \cite{loquercio_deep_2019, kaufmann_deep_2018, r_penicka_learning_2022, loquercio_learning_2021, y_song_learning_2023}.
Hybrid policies \cite{loquercio_learning_2021, r_penicka_learning_2022} demonstrate lower computational duration and higher success rates than traditional techniques. For instance, Agile \cite{loquercio_learning_2021} first eliminates mapping and planning modules by imitating actions provided by an expert with access to privileged information, thereby achieving autonomous flight at 10m/s in wild and human-made environments. Nevertheless, models trained through imitation learning often suffer from limited generalization due to their dependence on external guidance.
Therefore, instead of providing predefined actions from finite datasets, unsupervised learning, including reinforcement learning (RL), has been leveraged to train quadrotors to learn autonomously. This approach addresses the generalization limitations of supervised learning while avoiding the extensive effort required for expert preparation. MAVRL \cite{yu_mavrl_2024} employs an LSTM and an autoencoder, enabling the agent to exhibit long-term planning ability and adapt its velocity according to obstacle density. 
YOPOv2 \cite{lu_yopov2-tracker_2025} splits the field of view (FOV) into several patches, predicting the best trajectories within each patch along with their evaluated scores.
Beyond the aforementioned works, several methods \cite{kulkarni_reinforcement_2024, bhattacharya_vision_2025, kim_rapid_2025} directly map sensory inputs to control commands.

However, these methods still face challenges with onboard computation due to the increasing complexity of neural networks. The overall trend in collision-free algorithms is to introduce advanced learning techniques or backbones to improve performance, which inevitably increases the computational burden. Therefore, balancing performance and computation cost remains an open problem for collision-free flight.
Furthermore, most vision-based agile flight methods adopt acceleration \cite{yu_mavrl_2024}, velocity \cite{zhang_back_2024, bhattacharya_vision_2025}, or trajectory \cite{b_zhou_robust_2019, zhou_ego-planner_2021, kim_rapid_2025, lu_yopov2-tracker_2025, loquercio_learning_2021} as the command. Such approaches require an additional controller to track the command, which may fail to precisely follow the desired path during agile flight, as discussed in Section \ref{sec:introduction}.

\subsection{Reinforcement Learning via Differentiable Physics}
Reinforcement learning (RL) has shown great potential in solving aerial robot control tasks \cite{kaufmann_champion-level_2023, song_reaching_2023}, surpassing human performance in racing competitions. In typical RL frameworks, a critic estimates the long-term return, and an actor is trained to maximize the value output by the critic. Current RL methods can be categorized into model-free and model-based approaches. Model-free methods \cite{schulman2017proximal, haarnoja2018soft} directly learn the policy from interactions with the environment, which usually requires large amounts of data and a time-consuming training process. Model-based methods not only learn the policy from interactions but also build a transition model to predict future states. This transition model can be used for resampling imagined trajectories to improve data efficiency \cite{sutton1990integrated, janner2019trust}, serving as a fast-inferencing simulator to generate data for optimization \cite{chua2018deep, nagabandi2018neural}, or imagining future trajectories in a temporal latent space and then optimizing the policy through backpropagation \cite{hafner2019dream}. However, the difficulty of learning an accurate transition model or relying on a standalone critic to evaluate long-term returns limits the performance of traditional RL. If the optimization objective cannot be precisely estimated by critic, it significantly degrades actor training as well. 

In robotics, the transition model is inherently fully differentiable according to Newtonian physics, which enables the computation of precise desired gradients rather than relying on approximations from a critic. By replacing the critic with an analytical optimization objective, the actor achieves faster and more optimal convergence through a rollout of trajectories using backpropagation-through-time. Researchers have proposed a series of variants \cite{zhang_adaptive_2023, suh_differentiable_2022, mora_pods_2021} to improve stability and performance. Among them, the most recognized \cite{suh_differentiable_2022}, SHAC, splits long rollouts into short segments to avoid vanishing or exploding gradients and adds a value at the end of each segment to compensate for long-term returns. However, such algorithms have been primarily studied in simulation benchmarks over the past decade. Recently, differentiable physics has been successfully applied to aerial robots. For instance, Li \cite{li_visfly_2024} released the first differentiable simulator for quadrotors and successfully deployed it on several quadrotor benchmark tasks \cite{li2025abpt}. Xing \cite{pan2026learning} leveraged the superior convergence ability of differentiable physics to optimize policies online in the real world, addressing the immeasurable sim-to-real gap.
For vision-based tasks, Zhang \cite{zhang_back_2024} leverages thought of differentiable physics to train an obstacle-avoidance policy that enables a quadrotor to fly through dense forests at high speed without odometry. Although this work uses a point-mass model as the dynamics, it still demonstrates the potential of differentiable physics in vision-based agile flight.

\section{Method}
Previous works on collision avoidance have focused on simplifying or decoupling the overall problem—for example, by creating additional simplified environments specialized for training \cite{zhang_back_2024} or by inferring trajectories for a controller to track. Such approaches sacrifice dynamics fidelity for ease of training, thereby introducing an additional \textbf{training‑to‑inferring} gap beyond the sim‑to‑real gap. As a result, the learned policy must be validated in a high‑fidelity simulator before real‑world deployment, and it is nearly impossible to achieve the same performance as observed in the training environment due to tracking quality.

An overview of this research is shown in \cref{fig:overview}. In this work, we first align the control response from the real world to the simulation dimension by dimension. We then formulate the training as a reinforcement learning problem, relying directly on high‑fidelity differentiable simulation. This allows us to train the policy in the same environment, eliminating the need to create a separate training environment and thus removing the associated training‑to‑inferring gap. We address the sampling imbalance problem and local minima through distance‑based optimization. All frameworks are implemented following the standard architectures in Stable‑Baselines3 \cite{raffin2021stable}, making it easier for future research to build upon our work and to compare with other methods.

\subsection{RL preliminaries}

Reinforcement learning (RL) is a framework for sequential decision-making problems in which an agent learns to make decisions by interacting with an environment. It formulates the problem as a Markov Decision Process (MDP), defined by the tuple $(s_i, a_i, r_i, d_i, s_{i+1})$, where $s_i$ represents the state of the environment at time step $i$, $a_i$ is the action taken by the agent, $r_i$ is the reward received after taking action $a_i$, $d_i$ is a binary indicator of whether the episode has terminated, and $s_{i+1}$ is the next state. The most widely recognized architecture in RL is the actor–critic framework, which consists of two main components: the actor $\pi_{\theta}$, responsible for selecting actions based on the current observation, and the critic $Q_{\phi}$, which evaluates the long-term return of the current state–action pair.

In actor–critic methods, the actor is updated to maximize the value estimated by the critic. Depending on how the gradient of this objective is formed, gradient computation can be divided into policy‑gradient and value‑gradient methods.

Assuming a stochastic policy and trajectories sampled from that policy,
\begin{equation}
a \sim \pi_\theta(\cdot|s), \quad \tau \sim \pi_\theta,
\end{equation}
policy‑gradient methods estimate $\nabla_\theta J(\theta)$ from sampled trajectories using the likelihood‑ratio (score‑function) identity:
\begin{equation}
\nabla_\theta J(\theta)
=
\mathbb{E}{\tau \sim \pi\theta}
\left[
\sum_{t=0}^{T}
\nabla_\theta \log \pi_\theta(a_t|s_t),
A^{\pi_\theta}(s_t,a_t)
\right],
\end{equation}
where $\tau={(s_t,a_t,r_t)}{t=0}^{T}$ denotes a rollout generated by $\pi\theta$, $T$ is the rollout horizon (episode length), and $A^{\pi_\theta}(s_t,a_t)$ is the advantage function.

In contrast, value‑gradient methods update the actor by maximizing the critic's output via the action‑gradient of the critic and the chain rule. Specifically, the gradient can be expressed as
\begin{equation}
\nabla_\theta J(\theta)
= \mathbb{E}{s \sim \mathcal{D}}!\left[
\nabla\theta \pi_\theta(s),\nabla_a Q_\phi(s,a)
\right],
\end{equation}
where $\mathcal{D}$ denotes the empirical replay buffer used in off‑policy training.

Both policy-gradient and value-gradient methods rely on the quality of the critic $Q_\phi$ to provide accurate estimates of the expected return; otherwise, the actor may be misled to update in suboptimal directions. However, in environments with complex action and reward spaces, learning an accurate critic is particularly challenging, leading to high variance in policy updates and suboptimal performance. To address this challenge, differentiable simulation offers a promising solution by providing analytical gradients through the environment dynamics, thereby enabling faster and more stable convergence.

\subsection{Differentiable Physics and Backpropagation-Through-Time}
Unlike a learned critic that approximates long-term returns from sampled trajectories, differentiable physics provides an analytical reward function \(R(s_t,a_t)\) together with a differentiable state-transition model. Consider trajectories \(\tau=(s_0,a_0,\dots,s_{T},a_{T},s_{T+1})\) generated by rolling out a differentiable policy \(\pi_\theta\) over a horizon of length \(T\). The discounted cumulative reward to be optimized is
\begin{equation}
J(\theta)=\mathbb{E}_{\tau\sim\pi_\theta}\!\left[\sum_{t=1}^{T}\gamma^t R(s_t,a_t)\right],
\end{equation}
where \(\gamma\in(0,1]\) is the discount factor. Under the standard pathwise (reparameterization) assumption for differentiable rollouts, the gradient can be written as
\begin{equation}
\label{eq:grad}
\nabla_\theta J(\theta)
=\mathbb{E}_{\tau\sim\pi_\theta}\!\left[\sum_{t=1}^{T}\gamma^t \frac{dR(s_t,a_t)}{d\theta}\right].
\end{equation}
The derivative \(\frac{dR(s_t,a_t)}{d\theta}\) can be decomposed into the sum of two terms:
\begin{equation}
\label{eq:grad_split}
\frac{d R(s_t,a_t)}{d \theta}
=
\frac{\partial R(s_t,a_t)}{\partial s_t}\frac{\partial s_t}{\partial \theta}
+
\frac{\partial R(s_t,a_t)}{\partial a_t}\frac{\partial a_t}{\partial \theta}.
\end{equation}

The partial derivatives \(\partial R(s_t,a_t)/\partial s_t\) and \(\partial R(s_t,a_t)/\partial a_t\) are easily derived, as they are defined in the reward introduction (\cref{sec:reward_obs}). The action partial derivative \(\partial a_t / \partial \theta\) is directly related to the policy parameter \(\theta\). Considering that the current state evolves from the previous state and action, the state partial derivative \(\partial s_t / \partial \theta\) involves the entire past trajectory and can be expressed recursively as:
\begin{equation}
\frac{\partial s_t}{\partial \theta}
=\sum_{i=1}^t \left [ \left( \prod_{j=i+1}^t \frac{\partial s_j}{\partial s_{j-1}} \right) \frac{\partial s_i}{\partial a_{i-1}}\frac{\partial a_{i-1}}{\partial \theta} \right ],
\end{equation}
where \(\frac{\partial s_j}{\partial s_{j-1}}\) and \(\frac{\partial s_i}{\partial a_{i-1}}\) denote, respectively, the Jacobian of the dynamics and the Jacobian of the state transition with respect to the action, as will be introduced in \cref{sec:dynamics}.

Substituting the decomposition into the gradient expression, the analytical gradient of the actor in \cref{eq:grad} can finally be written as:
\begin{equation}
\begin{aligned}
\nabla_\theta J(\theta)
&=\mathbb{E}_{\tau\sim\pi_\theta}\!\Bigg [\sum_{t=1}^{T}\gamma^t \Bigg( \frac{\partial R(s_t,a_t)}{\partial a_t}\frac{\partial a_t}{\partial \theta} \\
&\qquad + \frac{\partial R(s_t,a_t)}{\partial s_t}\sum_{i=1}^t \Big [ \Big( \prod_{j=i+1}^t \frac{\partial s_j}{\partial s_{j-1}} \Big) \frac{\partial s_i}{\partial a_{i-1}}\frac{\partial a_{i-1}}{\partial \theta} \Big ] \Bigg)\Bigg ].
\end{aligned}
\end{equation}

\subsection{High-fidelity Dynamics of Quadrotors}
\label{sec:dynamics}

\begin{figure*}
    \centering
    \includegraphics[width=1.0\textwidth]{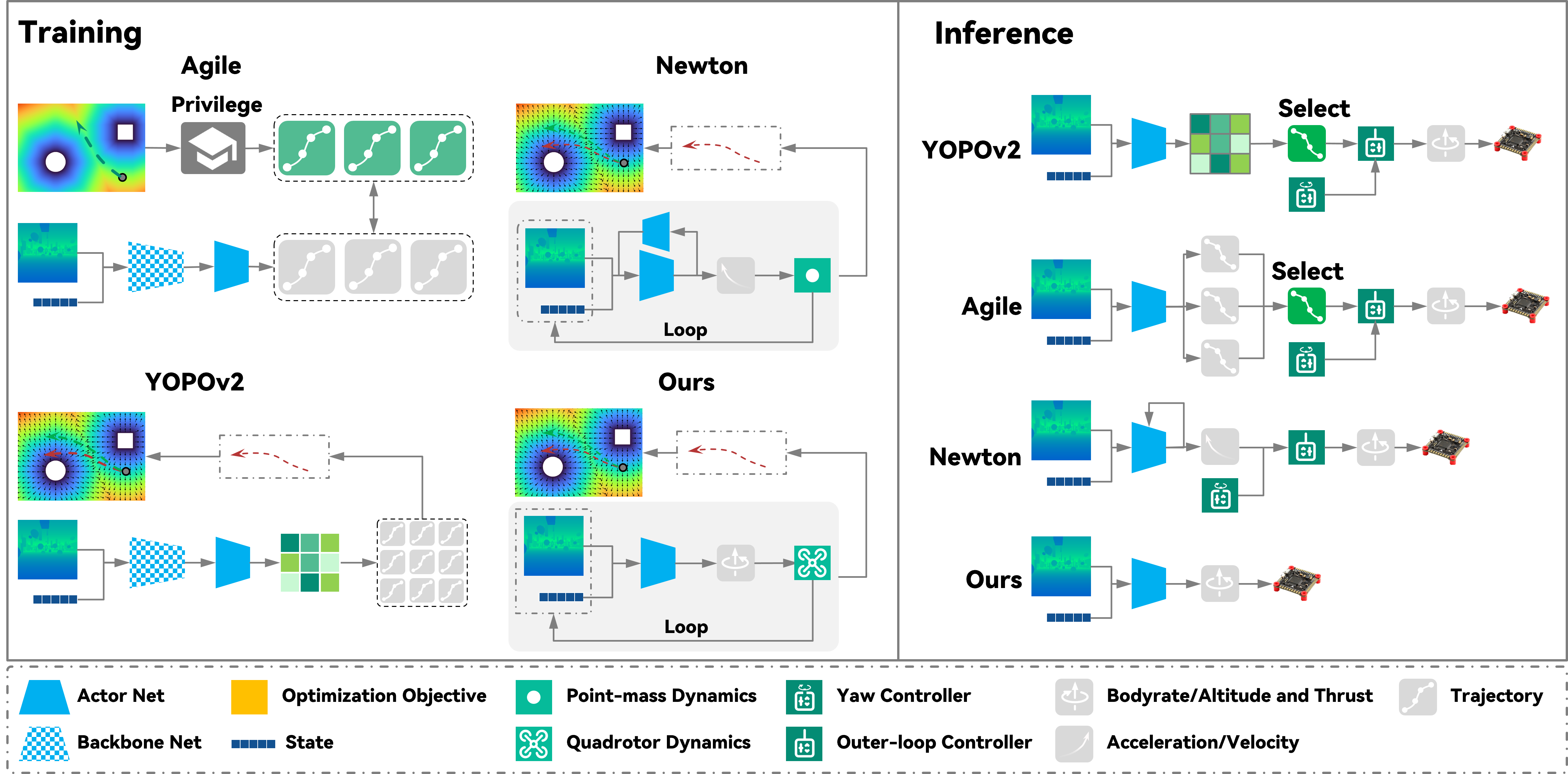}
    \caption{Training and inference pipeline of a state-of-the-art learning-based algorithm for collision-free flight without mapping. Our algorithm achieves true end-to-end control by eliminating intermediate trajectory representations in the reference frame. Direct generation of CTBR commands yields the shortest and most efficient inference chain, thereby avoiding dynamics-infeasible issues while improving maneuverability and stability.}
    \label{fig:enter-label}
\end{figure*}

Previous work \cite{zhang_back_2024} leveraging differentiable physics only employs a simple point-mass kinematics model to replace the complex quadrotor dynamics, which significantly reduces training complexity. However, this simplification sacrifices policy fidelity, leading to a \textbf{training-to-inferring gap} and \textbf{limiting the quadrotor's agility} due to the use of high-level commands. A more detailed illustration of the difference between our high-fidelity dynamics and prior kinematics is shown in \cref{fig:overview}.

In this work, we incorporate high-fidelity quadrotor dynamics directly into the gradient graph to better leverage agility while mitigating the training-to-deployment gap. In a precise simulation, the state $s=\{\mathbf{p},\mathbf{q}, \mathbf{v}, \mathbf{\Omega}\}$ of the quadrotor consists of position, orientation, linear velocity, and angular velocity. The CTBR action $a=\{\mathbf{\Omega}^d, {f}^d\}$ comprises desired body rates and collective thrust. Assuming a dynamics function $F$:
\begin{equation}
    \dot{s}=F(s, a)
\end{equation}
For actual code implementation, the quadrotor dynamics is usually formulated as a \textbf{time-discrete system}, thus transitioning to the next state as:
\begin{equation}
s_{t+1} = F(s_t, a_t) \, dt + s_t
\end{equation}
which makes it straightforward to obtain the partial derivatives $\frac{\partial s_{t+1}}{\partial s_t}$ and $\frac{\partial s_{t+1}}{\partial a_t}$ for backpropagation-through-time (BPTT):
\begin{equation}
    \begin{aligned}
        \frac{\partial s_{t+1}}{\partial s_t} &= \frac{\partial F}{\partial s_t} \, dt + I \\
        \frac{\partial s_{t+1}}{\partial a_t} &= \frac{\partial F}{\partial a_t} \, dt
    \end{aligned}
\end{equation}

The function $F$ can be defined using a widely adopted quadrotor autonomous ordinary differential equation:
\begin{equation}
\small
\label{eq:dynamics}
\begin{aligned}
    &\dot{\mathbf{p}}_W = {\mathbf{v}}_W \\
    &\dot{\mathbf{v}}_W = \frac{1}{m} \mathbf{R}_{WB}(\mathbf{f}  + \mathbf{d}) + \mathbf{g} \\
    &\dot{\mathbf{q}} = \frac{1}{2} \mathbf{q} \otimes \mathbf{\Omega} \\
    &\dot{\mathbf{\Omega}} = \mathbf{J}^{-1} \left (\boldsymbol{\eta} - \mathbf{\Omega} \times \mathbf{J} \mathbf{\Omega} \right )
\end{aligned}
\end{equation}
where $\mathbf{R}$ is the rotation matrix from the body frame to the world frame, $\boldsymbol{\eta}$ is the torque applied to the quadrotor, $\mathbf{d}$ is the disturbance force in the world frame, $m$ is the mass of the quadrotor, $\mathbf{J}$ is the inertia matrix, and $\mathbf{g}$ is the gravity vector. The operator $\otimes$ denotes quaternion multiplication, and $\mathbf{f}$ is the motor thrust in the body frame. For quadrotors, $\mathbf{f}$ can also be represented as $[0,0,f]^T$, where $f$ is the collective thrust.

Given the action, which includes desired body rates and collective thrust, we compute the desired torque $\boldsymbol{\eta}^d$ using a PD controller:
\begin{equation}
    \boldsymbol{\eta} = \mathbf{K}_p (\mathbf{\Omega}^d - \mathbf{\Omega}) + \mathbf{K}_d (\dot{\mathbf{\Omega}}^d - \dot{\mathbf{\Omega}})
\end{equation}
where $\mathbf{K}_p$ and $\mathbf{K}_d$ are the proportional and derivative gain matrices, respectively. The torque and collective thrust can be further mapped to individual motor thrusts using the quadrotor's configuration matrix:
\begin{equation}
    \label{eq:control_allocation}
    \begin{bmatrix}
        f_1 \\ f_2 \\ f_3 \\ f_4
    \end{bmatrix} = 
    \begin{bmatrix}
        1 & 1 & 1 & 1 \\
        0 & l & 0 & -l \\
        -l & 0 & l & 0 \\
        c & -c & c & -c
    \end{bmatrix}^{-1}
    \begin{bmatrix}
        f \\ \eta_x \\ \eta_y \\ \eta_z
    \end{bmatrix}
\end{equation}
where $l$ is the arm length from the motor to the center of mass, and $c$ is the induced torque coefficient. The terms $f_1, f_2, f_3, f_4$ are the individual motor thrusts, and $\eta_x, \eta_y, \eta_z$ are the torque components along the body-fixed $x$, $y$, and $z$ axes, respectively.

We use a second-order model to simulate the motor thrust as a function of rotor speed:
\begin{equation}
    \label{eq:motor_thrust}
    f_i = k_2 \omega_i^2 + k_1 \dot{\omega}_i + k_0
\end{equation}
where $k_2$, $k_1$, and $k_0$ are thrust coefficients estimated on a static thrust stand, and $\omega_i$ is the rotor speed of motor $i$. After obtaining the individual motor thrusts, the desired rotor speeds $\omega_i^d$ can be computed by inverting the second-order model:
\begin{equation}
    \omega_i^d = \sqrt{\frac{f_i - k_1 \dot{\omega}_i - k_0}{k_2}}
\end{equation}
To account for motor lag, the actuator response is approximated by a first-order inertial model:
\begin{equation}
    \omega_{i} = \omega_i + (\omega_i^d - \omega_i) \, e^{-dt / \tau}
\end{equation}
where $\tau$ is the motor time constant. Finally, the motor thrusts are recomputed using the actual rotor speeds via \cref{eq:motor_thrust}, and the collective external force and torque applied to the quadrotor are obtained inversely according to \cref{eq:motor_thrust,eq:control_allocation} for use in \cref{eq:dynamics} to compute the state derivatives.

Additionally, we simulate communication delay by maintaining a buffer of previous motor commands and applying them after a fixed delay duration:
\begin{equation}
    a_t = a_{t-D}
\end{equation}
where $D$ is the number of time steps corresponding to the communication delay.

Although incorporating the full quadrotor dynamics increases the complexity of the gradient graph, we find that it does not lead to gradient explosion or vanishing issues during BPTT within limited horizons. Despite the mathematical derivations being more complex, the implementation is straightforward and clear using automatic differentiation frameworks in PyTorch, as can be seen in our open-sourced code. Users need only override the reward function and the observation function when defining customized environment.

\subsection{Reward Function and Observation}
\label{sec:reward_obs}
We design the reward function to encourage the quadrotor to fly safely while minimizing control effort and ensuring smooth flight. The reward function comprises five components:
\begin{equation}
        r = c_v r_\mathbf{v} + c_p r_{p} + c_s r_s + c_a r_a + c_c r_c
\end{equation}
where the components are defined as follows.

A velocity tracking reward encourages the quadrotor to follow the desired velocity $\mathbf{v}^d$:
\begin{equation}
        r_\mathbf{v} = \mathrm{SmoothL1}(\|\mathbf{v}-\mathbf{v}^d\|, 0)
\end{equation}
where $\mathrm{SmoothL1}$ is the Huber loss function, which tolerates limited velocity errors caused by maneuvering to avoid obstacles.

A perception-aware reward is defined as the projection of the current velocity onto the camera viewing direction. Assuming the camera is rigidly mounted and aligned with the body frame's $x$-axis $\mathbf{x}_B$, this reward aligns the quadrotor's heading with its current velocity direction:
\begin{equation}
r_{p} =\frac{\mathbf{v} \cdot \mathbf{x}_B}{\|\mathbf{v}\| \|\mathbf{x}_B\|}
\end{equation}

A stability reward smooths the trajectory and penalizes aggressive maneuvers:
\begin{equation}
        r_s = \|\mathbf{\Omega}\| + c_{s2} \|\dot{\mathbf{v}}\| 
\end{equation}

A collision-avoidance penalty encourages the quadrotor to maintain a safe distance from obstacles:
\begin{equation}
r_a  =- \mathrm{clip}(d_{risk}-d_{col}-r_{uav}, 0)^2 \cdot
\mathrm{detach}(v_{appr}) 
\end{equation}
where $\mathrm{clip}$ bounds the value within a specified range, $d_{risk}$ is the distance threshold at which penalization begins, $d_{col}=\mathbf{p}-\mathbf{p}_{col}$ is the distance to the nearest obstacle, $r_{uav}$ is the radius of the quadrotor, and $v_{appr}$ is the approaching velocity toward the obstacle. The function $\mathrm{detach}$ removes the velocity from the gradient graph.

A collision penalty strongly discourages actual collisions:
\begin{equation}
  r_c = -32 \ln\left(e^{-7.5(d_{col}-r_{uav})}\right)
\end{equation}
We have also tested an inverse power function as the collision penalty, and this type of reward also performs well.

Note that only the position $\mathbf{p}$ within the rollout is backpropagated through the differentiable simulator to compute gradients; all other states are treated as ordinary variables during backpropagation. In our experiments, we found that for collision-free tasks, including velocity in the reward does not provide additional benefits.

\begin{figure*}
    \centering
    \includegraphics[width=1.0\linewidth]{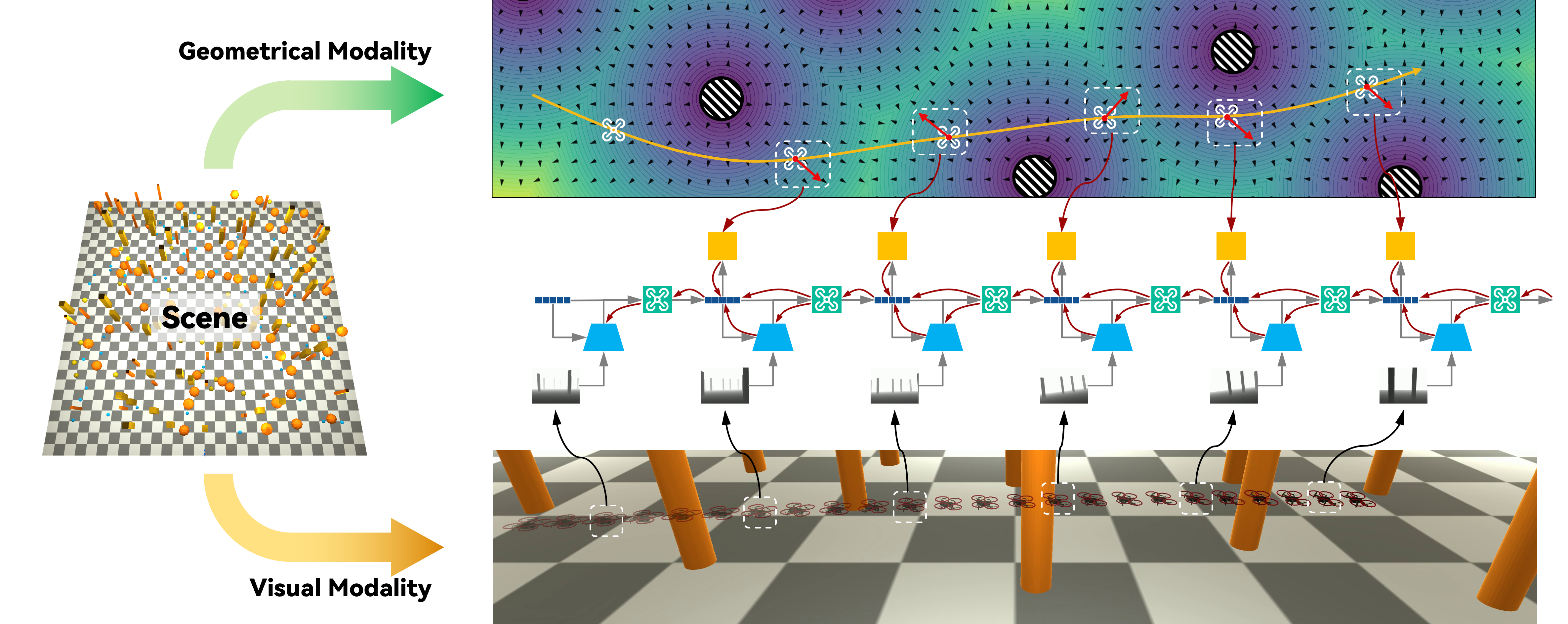}
    \caption{Multimodal representation supervision. Geometrical representation of one scene supervises the trajectory inferred by graphical representation. The gradient chain of differentiable simulation backpropagates the regulatory effect of the collision-based analytical reward.}
    \label{fig:multimodal}
\end{figure*}

This approach of using a differentiable collision reward to train the policy can be viewed as a form of multi-modal supervision within a temporal sequence. As shown in \cref{fig:multimodal}, a scene projects its depth features onto the camera image plane. The policy infers an action from the depth image, generating a temporal trajectory. This trajectory is supervised by another geometric representation of the same scene in the form of the collision reward. The safety regulation provided by this analytical reward is backpropagated through the rollout, enabling end-to-end policy training in a single step.

The policy observes the current state, the desired velocity $\mathbf{v}^d$, and the depth image, then outputs CTBR commands. To reduce the observation space, we remap the desired velocity $\mathbf{v}^d$ and the current velocity $\mathbf{v}$ into the heading frame of the quadrotor:
\begin{equation}
\mathbf{v}_H = \mathbf{R}_{WH} \mathbf{v}, \quad
\mathbf{v}^d_H = \mathbf{R}_{WH} \mathbf{v}^d
\end{equation}
where $\mathbf{v}_H$ and $\mathbf{v}^d_H$ are the current and desired velocities in the heading frame, respectively, and $\mathbf{R}_{WH}$ is the transformation matrix from the world frame to the heading frame. The heading frame is obtained from the world frame by rotating only the yaw angle, so $\mathbf{R}_{WH}$ is defined as:
\begin{equation}
\mathbf{R}_{WH} = 
\begin{bmatrix}
\cos\psi & \sin\psi & 0 \\
-\sin\psi & \cos\psi & 0 \\
0 & 0 & 1
\end{bmatrix}
\end{equation}

The depth image is preprocessed by inverting the values and applying max-pooling to reduce it to a resolution of $12 \times 16$, which reduces the observation dimension while focusing on the closest obstacles. The final observation consists of two parts: (1) a concatenation of the desired velocity in the heading frame $\mathbf{v}^d_H$, the current velocity in the heading frame $\mathbf{v}_H$, the angular velocity $\mathbf{\Omega}$, and the orientation $\mathbf{q}$ as a quaternion; and (2) the preprocessed depth image.

\subsection{Gradient Field Reshaping}
\begin{figure*}
    \centering
    \includegraphics[width=1.0\linewidth]{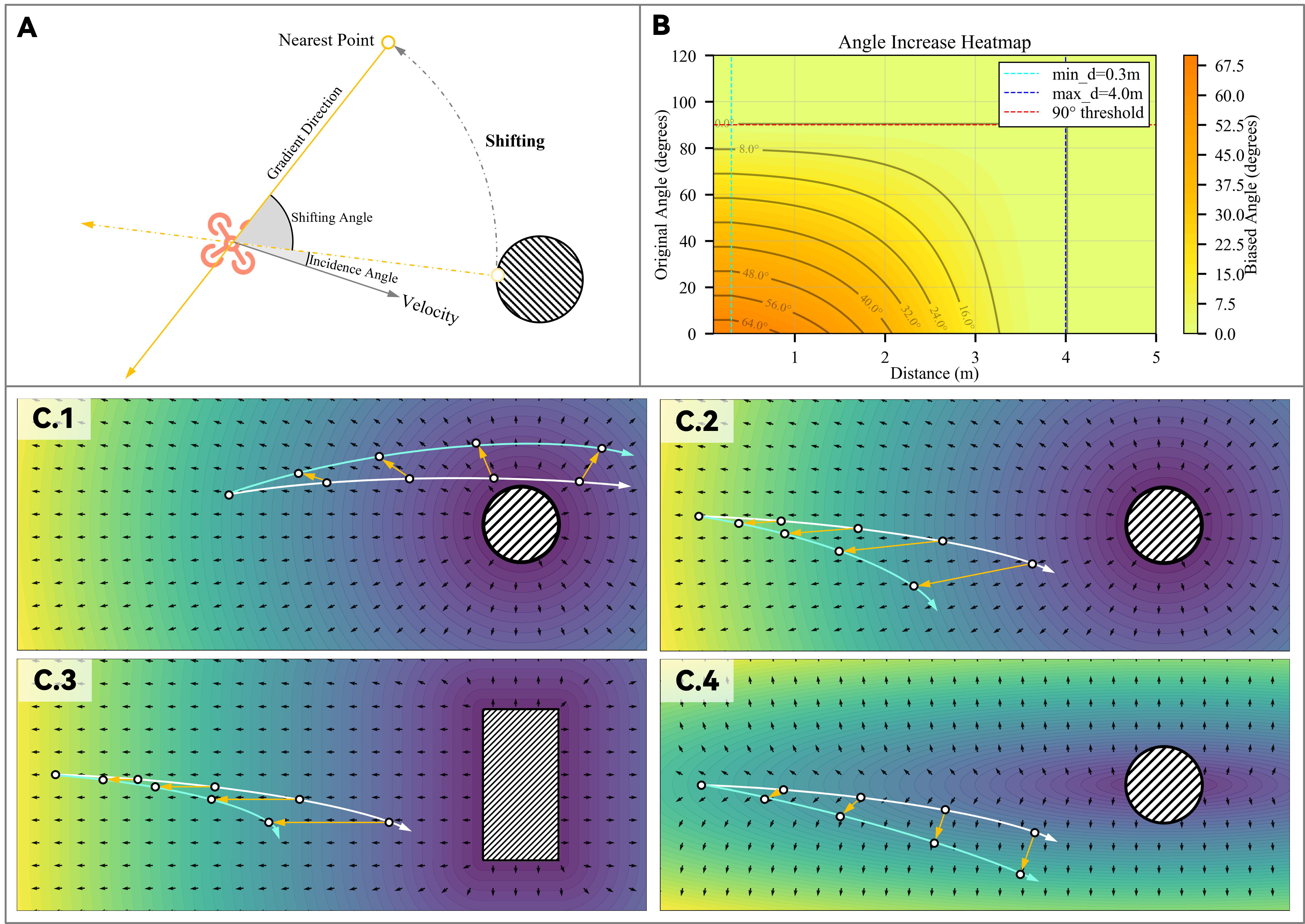}
    \caption{A schematic diagram of ESDF map reshaping. \textbf{(A)}: The shifting operation applied to the nearest obstacle point, which transforms the gradient direction from opposing the velocity to the lateral direction. \textbf{(B)}: The shifting angle $\Delta a$ as a function of distance to the nearest obstacle $d$ and incidence angle $a$. \textbf{(C.1 and C.2)}: The ESDF gradient field for scenes containing one round obstacle and one rectangular obstacle, where the "death zone" with nearly uniform gradient direction is highlighted in red. \textbf{(C.3)}: An approximate visualization of the reshaped gradient field. \textbf{(C.4)} The gradient field after reshaping. Note that the gradient fields in \textbf{(C)} are not exact ESDFs; they are the same only without the shifting operation.}
    \label{fig:map_reshape}
\end{figure*}

When maximizing the reward, the nearest collision distance requires a gradient, consistent with previous optimization-based methods \cite{zhou_ego-swarm_2021}. Consequently, this regulatory effect reflected in the gradient field is equivalent to that of the Euclidean Signed Distance Field (ESDF), which represents the relative distance to the nearest obstacle at each point in space.

However, such optimization methods naturally tend to get stuck in local minima during high-speed flight, leading to suboptimal policies. Specifically, as shown in \cref{fig:map_reshape} (C.1), the ESDF gradient points in the direction opposite the nearest obstacle, biasing the trajectory to maintain a safe clearance. Yet, if the quadrotor's velocity is directly opposite the ESDF gradient direction, the policy tends to brake rather than avoid, which is undesirable for collision-free flight. From the perspective of gradient fields, the tendency to brake rather than avoid is proportional to the incidence angle between the velocity vector and the gradient line. When the velocity vector is parallel to the gradient line, the policy is more likely to brake; when the velocity vector is perpendicular to the gradient line, the policy is more likely to avoid. This issue becomes more severe under two circumstances: (1) when the quadrotor is flying at high speed, it fails to sense probable collisions far enough in advance within the same limited duration, and (2) when the scene contains large obstacles rather than thin trees (as in \cref{fig:map_reshape}.C.2), the gradient field contains large contiguous regions with nearly uniform gradient lines, making it difficult for quadrotors to make avoidance decisions along the foreseeable trajectory, trapping them in this "death zone".

To address this issue, we apply a transformation to the detected nearest obstacle point, shifting the optimization effect from braking to avoiding while preserving the desired velocity command. Specifically, as illustrated in \cref{fig:map_reshape}.A, the forward-pointing direction is shifted from forward to lateral by an angle that depends on the distance and the current incidence angle. This shifting operation transforms the gradient direction from opposing the velocity to the lateral direction, thereby training the policy to avoid rather than brake. The shifting angle $\Delta a$ is defined as a function of the distance to the nearest obstacle $d$ and the incidence angle $a$:
\begin{equation}
\begin{aligned}
\tilde d&=\mathrm{clip}(d,d_{\min},d_{\max}), \quad
x=\frac{\tilde d-d_{\min}}{d_{\max}-d_{\min}} \\
\tilde a&=\mathrm{clip}(a, 0, a_{\max}) \\
\Delta a&=(a_{\max}-\tilde a)\,\alpha\left[1-\exp({-3(1-x)^2})\right]
\end{aligned}
\end{equation}
where $d_{\min}$, $d_{\max}$, and $a_{\max}$ represent the minimum-distance, maximum-distance, and maximum-angle thresholds for shifting, respectively, and $\alpha$ is the maximum shifting angle. \cref{fig:map_reshape}.B illustrates the additional shifting as a function of incidence angle and distance. The shifting angle increases as the obstacle approaches and the incidence angle rises, indicating higher risk and a stronger tendency to avoid. The incidence-angle threshold $a_{\max}$ is set to $\pi/2$, since shifting is unnecessary when the quadrotor is flying away from or past the obstacle. This transformation removes the "death zone" from the gradient field: as shown in \cref{fig:map_reshape}.C.3, the gradient in front of the obstacle points laterally, thus optimizing the policy to fly around the obstacle.

\subsection{Training Settings}
\subsubsection{Network Architecture}

The actor and critic networks follow a feature-extractor–MLP design as in Stable-Baselines3 \cite{raffin2021stable}. The feature extractor consists of two branches: a three-layer CNN with channel sizes [32, 64, 128] for processing the depth image, and a two-layer MLP with hidden widths [512, 512] for the concatenated state vector. The extracted features are concatenated and passed to separate two-layer MLP heads, each with [512, 512] units, for the actor and critic. The actor outputs four-dimensional CTBR commands in the final linear layer, while the critic outputs a scalar Q-value. All layers use LeakyReLU activations.

To the best of our knowledge, this is the simplest architecture that can achieve high-speed flight in dense obstacle environments. Its clear and simple structure benefits real-world deployment, as it avoids the temporal instability associated with recurrent layers and respects the limited hardware resources on the quadrotor.

\subsubsection{Domain Randomization}

\begin{figure*}
    \centering
    \includegraphics[width=0.8\linewidth]{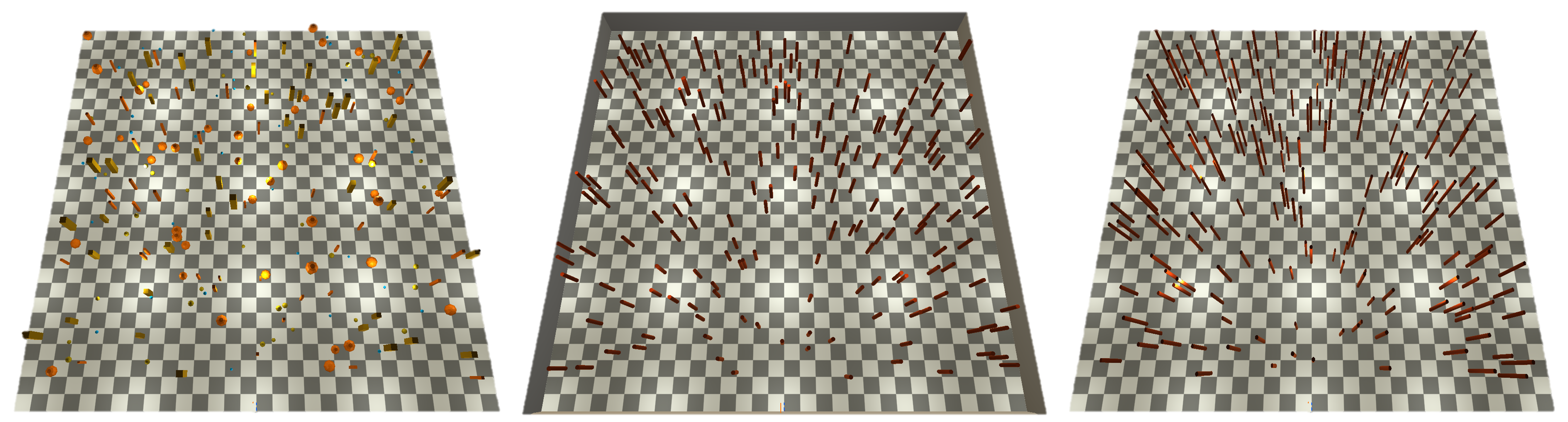}
    \caption{Representative examples of the three groups of training scenes. The first group consists of boxes and pillars of varying shapes. The second group comprises short pillars. The third group contains tall pillars.}
    \label{fig:train_scene}
\end{figure*}

To improve the robustness of the learned policy, we apply domain randomization during training. Because we have performed precise system identification of the quadrotor, we only randomize the environment properties and difficultly-recognized parameters like aerodynamic coefficients. We define three groups of scenes, as shown in \cref{fig:train_scene}, which include varying obstacle densities, shapes, and spatial distributions. Additionally, we randomize the initial state of the quadrotor and the desired velocity.

In contrast to specialized training environment surrogates, high-fidelity simulation introduces a systematic sampling bias in reinforcement learning. Specifically, when the desired speed \(v^d\) is sampled from a uniform distribution, slower trajectories persist longer in a finite rendered scene and thus contribute disproportionately more samples, skewing the dataset toward low-speed regimes. To mitigate this bias, we reshape the distribution of desired velocities so that the effective distribution of \(v^d\) in the collected data matches a target density.

Let \(p_{\text{cmd}}(v^d)\) denote the sampling density of the desired velocity and \(p_{\text{data}}(v^d)\) the effective density observed in the dataset. The desired velocity is randomized over \([v_{\min}, v_{\max}]\). If the collected data followed a uniform distribution over desired velocity, we would have \(p_{\text{data}}(v^d) = 1/(v_{\max}-v_{\min})\). Under the assumption that the number of samples collected from a trajectory is proportional to its duration, and that the duration \(t\) scales as \(t \propto 1/v^d\), the induced density satisfies
\begin{equation}
p_{\text{data}}(v^d) \propto \frac{1}{v^d}\,p_{\text{cmd}}(v^d).
\end{equation}
We therefore choose
\begin{equation}
p_{\text{cmd}}(v^d) \propto v^d \cdot \frac{1}{v_{\max} - v_{\min}}.
\end{equation}
After normalization, the sampling density of the desired velocity is given by
\begin{equation}
p_{\text{cmd}}(v^d) = \frac{2v^d}{v_{\max}^2 - v_{\min}^2}, \quad v^d \in [v_{\min}, v_{\max}].
\end{equation}
In practical implementation, to sample the desired velocity, we draw \(u \sim \mathcal{U}(0,1)\) and compute
\begin{equation}
v^d = \sqrt{v_{\min}^2 + u\,(v_{\max}^2 - v_{\min}^2)},
\end{equation}
which yields a uniform distribution of \(v^d\) in the collected data.

\section{Experiments}
We conduct extensive experiments to evaluate the performance of our proposed method against state-of-the-art algorithms in simulation. Based on the results, we provide a detailed comparison and analysis of the advantages of each method, along with the underlying reasons for their relative performance. Subsequently, we implement the proposed method on a real-world robotic system and conduct experiments to verify its generalization capability in outdoor scenarios. The experimental results demonstrate that our method achieves superior stability and agility compared to state-of-the-art approaches.

\subsection{Training Result}
\begin{figure}
    \centering
    \includegraphics[width=1.0\linewidth]{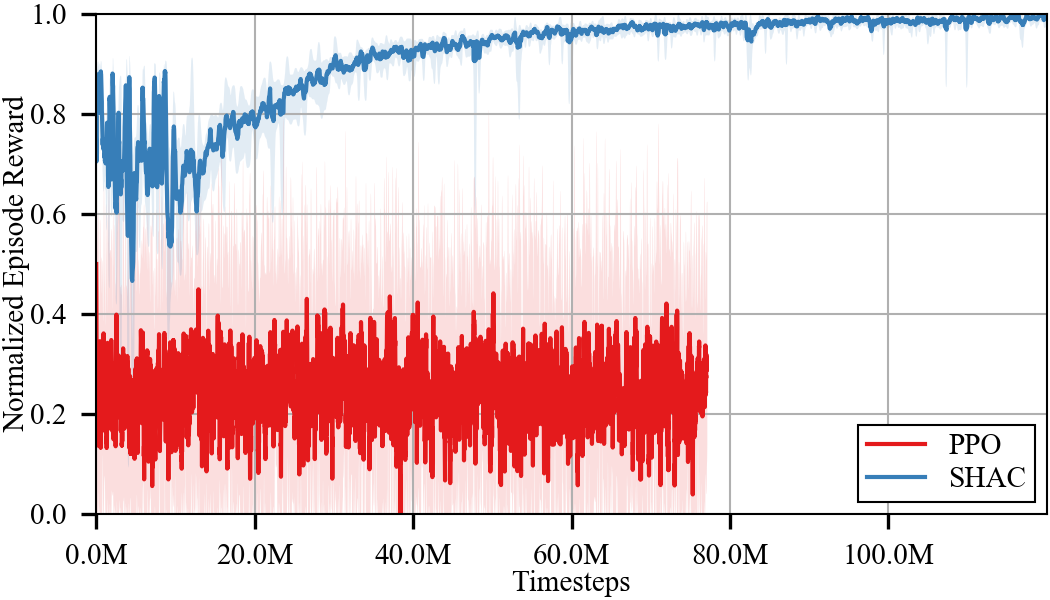}
    \caption{Normalized training curves of SHAC and PPO.}
    \label{fig:trainingRes}
\end{figure}

We train the policy using SHAC \cite{xu_accelerated_2022}—a variant of backpropagation-through-time (BPTT)—and implement the algorithm ourselves. To verify the necessity of using SHAC, we compare its training results with those of the popular reinforcement learning algorithm PPO \cite{schulman2017proximal}. The hyperparameters of both algorithms are carefully fine-tuned to achieve the best possible performance. The training results are shown in \cref{fig:trainingRes}. In contrast to \cite{zhang_back_2024}, where PPO converged to a slightly lower performance than BPTT, PPO fails to train the policy in our environment due to the increased complexity of the transition function. This result demonstrates that analytical gradients benefit robot learning in domains with complex robot system.

\subsection{Simulated Experiments}
\subsubsection{Experimental Setup and Baselines}

To compare all methods under a common perceptual environment, we evaluate them in the same visually rendered scenes provided by VisFly \cite{li_visfly_2024}. The rendered geometry, obstacle layouts, and depth observations are therefore shared across methods. In preliminary experiments, we found that several baselines were not robust to changes in the simulation backend, controller stack, or dynamics environment, and their performance degraded substantially when deployed outside the setup used in their original implementations. To provide the fairest comparison, we therefore evaluate each baseline in its native simulation stack—the environment in which it was originally deployed and fine-tuned—while using VisFly solely to provide the same rendered scene and depth input across all methods.

Our policy is evaluated directly within VisFly, which natively handles both the six-degree-of-freedom (6-DOF) rigid-body dynamics and the depth rendering pipeline in a unified simulator. For the baselines, VisFly serves only as the rendering engine, while each baseline's native simulator remains responsible for dynamics propagation, state estimation, and low-level control. During evaluation, the baseline simulator continuously sends the quadrotor's pose and odometry to VisFly via direct state assignment; VisFly then renders the corresponding depth image and returns it to the baseline algorithm. 

This setup allows every method to perceive the same scene while preserving the dynamics and controller stack under which each baseline achieves its best performance, as originally reported in its respective paper.

We compare our end-to-end policy against four representative baselines: EGO-Planner \cite{zhou_ego-planner_2021}, Agile \cite{loquercio_learning_2021}, Newton \cite{zhang_back_2024}, and YOPOv2 \cite{lu_yopov2-tracker_2025}. EGO-Planner is a modular sense–map–plan method. Agile is an imitation-learned policy distilled from expert behavior. Newton is a method leveraging a simplified kinematics model. YOPOv2 is a one-stage perception-driven planner based on motion primitives that also leverages differentiable physics for trajectory optimization. Together, these baselines span modular planning, imitation learning, and differentiable-simulation-based control.

\subsubsection{Comparison in Large-scale Map}
We evaluate all methods in a large-scale forest environment with four obstacle densities (0.02, 0.04, 0.06, and 0.08 obstacles/m²) and four commanded velocities (3, 6, 9, and 12 m/s). For each density, we generate five scene instances and evaluate five different start positions, yielding 25 runs per configuration.

\begin{figure*}
    \centering
    \includegraphics[width=1.0\textwidth]{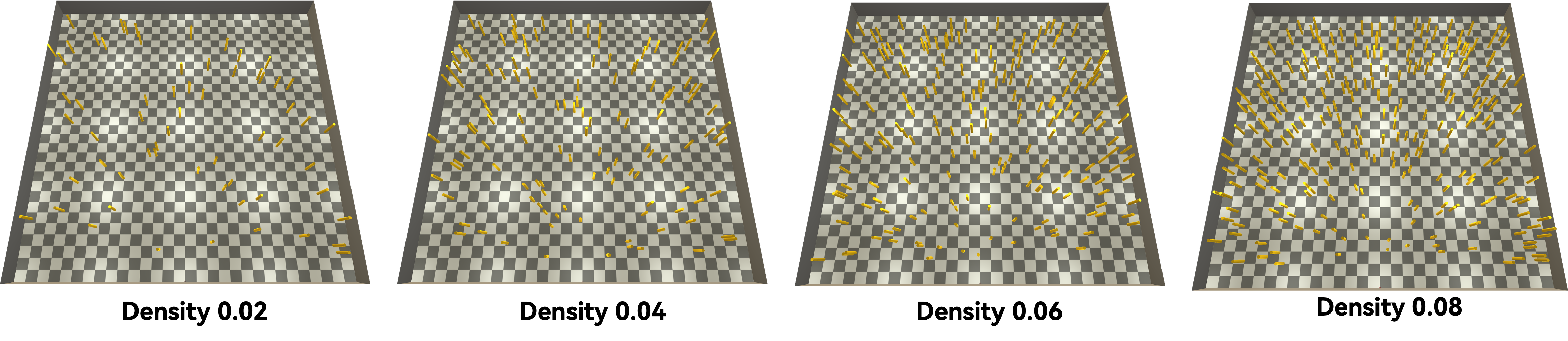}
    \caption{
        Large-scale forest environments at four obstacle densities: 0.02, 0.04, 0.06, and 0.08 obstacles/m$^2$. For each density, we generate five map instances and evaluate each instance from five different start positions as \cref{fig:performance_across_densities_speeds}.A, with the goal located on the opposite side of the forest.
    }
    \label{fig:large_scale_scenes}
\end{figure*}

\begin{figure*}
    \centering
    \includegraphics[width=1.0\textwidth]{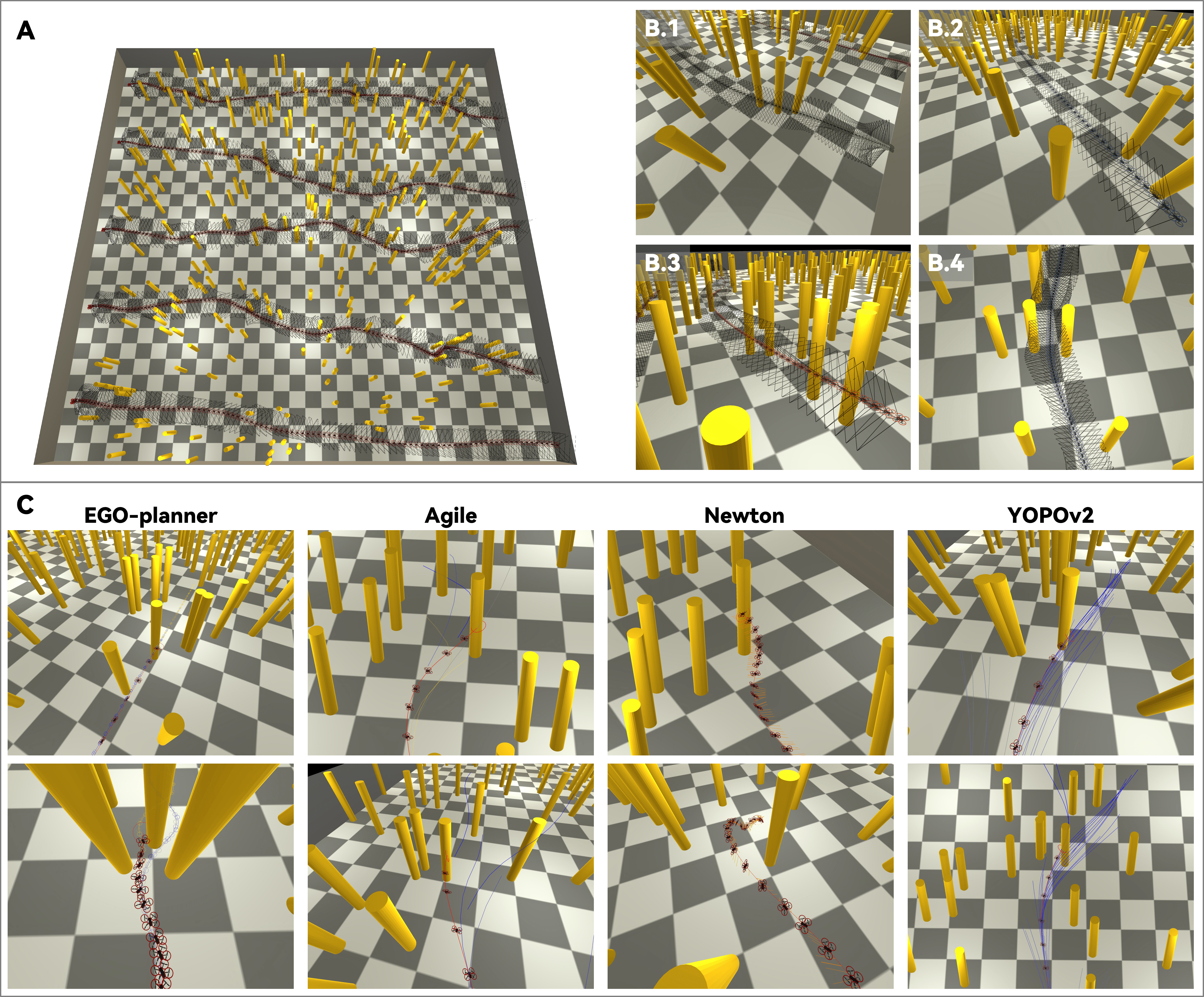}
    \caption{
        \textbf{(A)} Trajectories of five test cases in one scene. 
        \textbf{(B)} Trajectories of our policy.
        \textbf{(C)} Failure cases of baselines.
    }
    \label{fig:sim_comparison}
\end{figure*}

Representative scenes are shown in Figure~\ref{fig:large_scale_scenes}, and the quantitative results are summarized in Table~\ref{tab:comparision}. Figure~\ref{fig:sim_comparison} illustrates instances in which the baselines fail, providing insight into the underlying reasons for their performance degradation in such scenarios, as discussed in the following paragraph.

\begin{table*}[ht!]
\centering
\setlength{\tabcolsep}{3.0pt}
\renewcommand{\arraystretch}{1.15}
\caption{Key Character Comparison of Baselines across All Velocities and Densities. \label{tab:comparision}}
\begin{tabular*}{\textwidth}{@{\extracolsep{\fill}}l c | cccc | cccc | cccc | cccc@{}}
\toprule

\multirow{2}{*}{\textbf{Method}} & \multirow{2}{*}{\textbf{Dens.}} &
\multicolumn{4}{c|}{\textbf{v = 3 m/s}} &
\multicolumn{4}{c|}{\textbf{v = 6 m/s}} &
\multicolumn{4}{c|}{\textbf{v = 9 m/s}} &
\multicolumn{4}{c}{\textbf{v = 12 m/s}} \\
\cmidrule(lr){3-6} \cmidrule(lr){7-10} \cmidrule(lr){11-14} \cmidrule(lr){15-18}

& &
\textbf{S/T} & \textbf{Comp} & \textbf{Vel} & \textbf{Jerk} &
\textbf{S/T} & \textbf{Comp} & \textbf{Vel} & \textbf{Jerk} &
\textbf{S/T} & \textbf{Comp} & \textbf{Vel} & \textbf{Jerk} &
\textbf{S/T} & \textbf{Comp} & \textbf{Vel} & \textbf{Jerk} \\
\midrule

\multirow{4}{*}{\shortstack[l]{\textbf{EGO-}\\\textbf{Planner}}}
& 0.02 & \textbf{25/25} & \textbf{1.000} & 2.49 & 4.92 & 24/25 & 0.963 & 4.72 & 22.06 & 11/25 & 0.651 & 6.05 & 45.13 & 9/25 & 0.632 & 7.15 & 81.32 \\
& 0.04 & 24/25 & 0.963 & 2.48 & 9.23 & 18/25 & 0.863 & 4.64 & 26.08 & 8/25 & 0.601 & 5.79 & 50.78 & 4/25 & 0.570 & 7.07 & 79.58 \\
& 0.06 & 23/25 & 0.946 & 2.48 & 9.60 & 18/25 & 0.816 & 4.58 & 35.48 & 7/25 & 0.526 & 5.51 & 62.11 & 3/25 & 0.444 & 6.58 & 78.70 \\
& 0.08 & \textbf{25/25} & \textbf{1.000} & 2.47 & 11.43 & 12/25 & 0.737 & 4.51 & 45.23 & 4/25 & 0.498 & 5.64 & 64.68 & 0/25 & 0.370 & 6.67 & 80.89 \\
\midrule

\multirow{4}{*}{\textbf{Agile}}
& 0.02 & 17/25 & 0.854 & 2.96 & 18.44 & 15/25 & 0.815 & 5.39 & 52.26 & 13/25 & 0.776 & 6.88 & 124.22 & 12/25 & 0.754 & 7.65 & 218.38 \\
& 0.04 & 9/25  & 0.654 & 2.88 & 25.21 & 9/25  & 0.717 & 5.16 & 60.99 & 9/25  & 0.695 & 6.40 & 128.31 & 5/25  & 0.603 & 6.78 & 239.12 \\
& 0.06 & 8/25  & 0.631 & 2.88 & 28.01 & 7/25  & 0.556 & 4.74 & 67.88 & 5/25  & 0.520 & 5.78 & 135.63 & 0/25  & 0.549 & 6.35 & 215.80 \\
& 0.08 & 5/25  & 0.558 & 2.89 & 29.48 & 3/25  & 0.542 & 4.93 & 72.55 & 4/25  & 0.527 & 5.96 & 144.25 & 2/25  & 0.454 & 5.46 & 197.79 \\
\midrule

\multirow{4}{*}{\textbf{Newton}}
& 0.02 & \textbf{25/25} & \textbf{1.000} & 2.87 & 3.93 & \textbf{25/25} & \textbf{1.000} & 5.30 & 12.66 & \textbf{25/25} & \textbf{1.000} & 7.12 & 35.04 & \textbf{25/25} & \textbf{1.000} & 7.56 & 97.82 \\
& 0.04 & \textbf{25/25} & \textbf{1.000} & 2.86 & 6.24 & \textbf{25/25} & \textbf{1.000} & 5.15 & 19.74 & 23/25 & 0.982 & 6.50 & 64.53 & 22/25 & 0.944 & 6.59 & 135.87 \\
& 0.06 & \textbf{25/25} & \textbf{1.000} & 2.86 & 7.69 & \textbf{25/25} & \textbf{1.000} & 5.07 & 23.09 & 22/25 & 0.980 & 6.30 & 74.12 & \textbf{23/25} & \textbf{0.969} & 6.37 & 134.74 \\
& 0.08 & \textbf{25/25} & \textbf{1.000} & 2.80 & 9.11 & \textbf{25/25} & \textbf{1.000} & 4.86 & 29.69 & \textbf{23/25} & \textbf{0.979} & 5.84 & 67.66 & 19/25 & 0.920 & 5.59 & 112.26 \\
\midrule

\multirow{4}{*}{\textbf{YOPOv2}}
& 0.02 & \textbf{25/25} & \textbf{1.000} & 2.74 & 4.44 & \textbf{25/25} & \textbf{1.000} & 5.26 & 11.43 & 23/25 & 0.927 & 7.27 & 36.06 & 23/25 & 0.931 & \textbf{9.14} & 64.57 \\
& 0.04 & \textbf{25/25} & \textbf{1.000} & 2.70 & 4.38 & \textbf{25/25} & \textbf{1.000} & 5.18 & 12.81 & \textbf{24/25} & 0.967 & 7.42 & 31.43 & 18/25 & 0.835 & \textbf{8.72} & 66.49 \\
& 0.06 & \textbf{25/25} & \textbf{1.000} & 2.67 & \textbf{3.89} & \textbf{25/25} & \textbf{1.000} & 5.11 & 14.24 & \textbf{23/25} & 0.963 & \textbf{7.37} & 30.02 & 14/25 & 0.714 & 7.87 & 74.96 \\
& 0.08 & 22/25 & 0.957 & 2.58 & \textbf{3.76} & 23/25 & 0.987 & 5.01 & 14.77 & 21/25 & 0.864 & 6.75 & 43.05 & 12/25 & 0.750 & \textbf{8.41} & 66.76 \\
\midrule

\multirow{4}{*}{\textbf{Ours}}
& 0.02 & \textbf{25/25} & \textbf{1.000} & \textbf{3.68} & \textbf{2.97} & \textbf{25/25} & \textbf{1.000} & \textbf{5.76} & \textbf{5.52} & \textbf{25/25} & \textbf{1.000} & \textbf{7.66} & \textbf{9.56} & 24/25 & 0.984 & 8.97 & \textbf{15.72} \\
& 0.04 & \textbf{25/25} & \textbf{1.000} & \textbf{3.48} & \textbf{3.80} & \textbf{25/25} & \textbf{1.000} & \textbf{5.60} & \textbf{7.58} & \textbf{24/25} & \textbf{0.993} & \textbf{7.43} & \textbf{11.80} & \textbf{24/25} & \textbf{0.973} & 8.79 & \textbf{17.72} \\
& 0.06 & \textbf{25/25} & \textbf{1.000} & \textbf{3.45} & 4.04 & \textbf{25/25} & \textbf{1.000} & \textbf{5.40} & \textbf{9.29} & \textbf{23/25} & \textbf{0.962} & 7.08 & \textbf{14.80} & 22/25 & 0.938 & \textbf{8.05} & \textbf{21.71} \\
& 0.08 & \textbf{25/25} & \textbf{1.000} & \textbf{3.43} & 4.56 & \textbf{25/25} & \textbf{1.000} & \textbf{5.14} & \textbf{10.87} & \textbf{25/25} & \textbf{1.000} & \textbf{6.84} & \textbf{15.80} & \textbf{23/25} & \textbf{0.976} & 7.61 & \textbf{23.19} \\
\bottomrule
\end{tabular*}

\begin{tablenotes}
\item \textbf{S/T}: Success / Total Runs; \textbf{Comp}: Average Completion Rate; \textbf{Vel}: Average Velocity (m/s); \textbf{Jerk}: Average Jerk (m/s$^3$).
\item All velocity metrics, including average velocity and maximum velocity, are measured as the effective forward velocity along the goal direction, so detours around obstacles reduce the reported speed instead of inflating it.
\end{tablenotes}
\end{table*}

\begin{figure*}[htbp]
    \centering
    \includegraphics[width=1.0\textwidth]{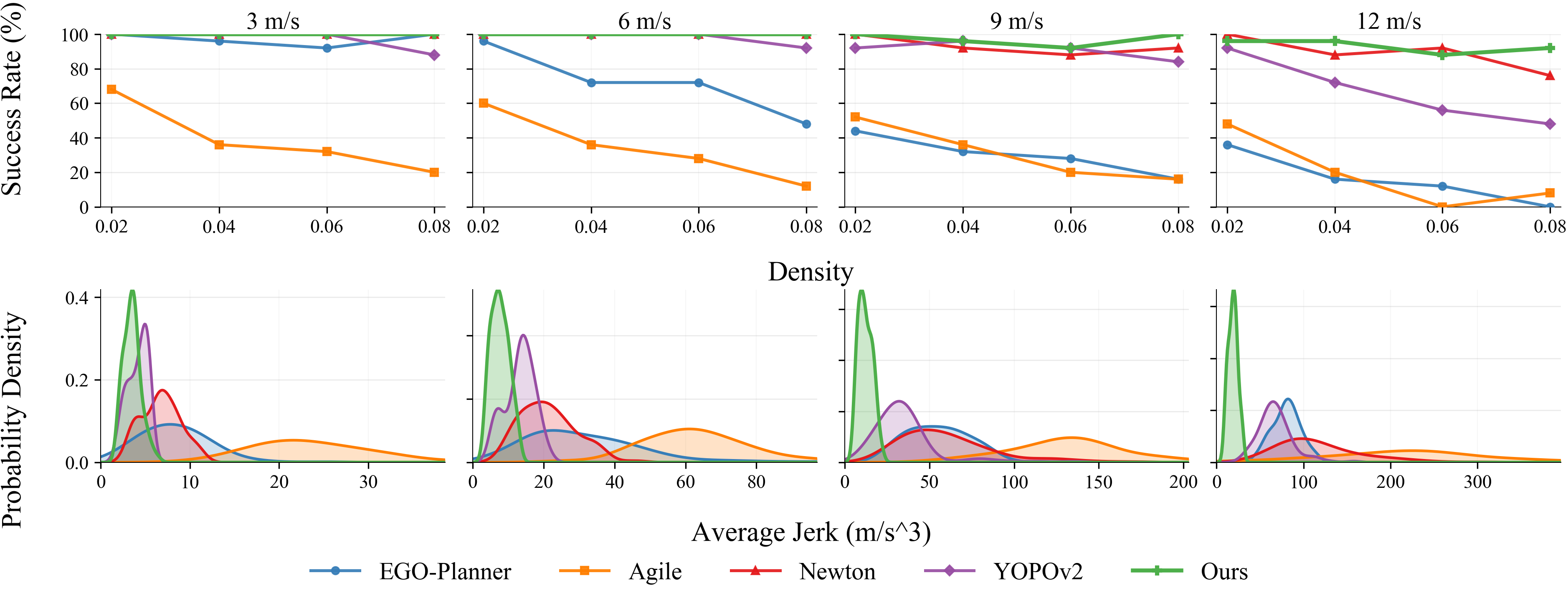}
    \caption{\textbf{Top:} Comparison of success rates under varying obstacle densities (0.02--0.08\,obs/m$^2$) and commanded speeds (3--12\,m/s). Our method maintains near-perfect reliability even in the most aggressive and dense environments. \textbf{Bottom:} Probability density distribution of average jerk across speeds. Our end-to-end policy executes the smoothest maneuvers with consistently lower overall jerk compared to all baselines.}
    \label{fig:performance_across_densities_speeds}
\end{figure*}

EGO-Planner performs strongly at low speeds, achieving near-perfect success rates at 3m/s across all tested densities. Its performance, however, drops sharply as the commanded speed increases, particularly in dense scenes. This degradation is consistent with a fundamental limitation of its modular pipeline: obstacle avoidance depends on a finite planning and replanning cycle, and at high speeds the quadrotor can travel a substantial distance before a new plan is generated. Consequently, the controller is often forced to execute a plan that is already outdated relative to the current scene, reducing the available margin for safe avoidance. This effect becomes most severe at 9–12m/s in cluttered environments, where the success rate falls to as low as 0/25. These results indicate that, for EGO-Planner, the main bottleneck in aggressive flight is not the nominal command itself but rather the latency of the planning pipeline. Furthermore, in real-world deployments, EGO-Planner requires constructing precise maps during flight, which is highly sensitive to odometry drift due to system noise. Such drift can compound mapping errors and exacerbate latencies, further degrading overall system performance.

Agile maintains moderate success at low speeds in sparse environments but degrades rapidly as both commanded speed and obstacle density increase. A likely reason is that its imitation-learned policy is closely tied to the state distribution of the expert demonstrations; consequently, in aggressive flight, even small errors can push the quadrotor into states from which recovery is more difficult. In our rollouts, this limitation manifests as clear instability: the vehicle often exhibits winding flight patterns, and in dense scenes it makes frequent alternating left–right avoidance turns instead of committing to a stable maneuver. We also observe high temporal variance in the predicted trajectory, with the avoidance direction sometimes changing abruptly between consecutive decisions. This directional inconsistency reduces effective forward velocity and leaves insufficient time for stable recovery near obstacles. Consistent with these observations, even when commanded to fly at 12m/s, Agile's best average forward velocity remains below 7.7m/s, while its jerk reaches 239.12m/s³, the highest among all evaluated methods. Taken together, these results suggest that Agile's primary limitation in this benchmark is the reduced robustness of its imitation-learned reactive policy under fast, cluttered flight.

Among all baselines, YOPOv2 achieves the highest average forward velocity and maintains perfect or near-perfect success rates at low to moderate speeds. However, its performance degrades notably at 12m/s in denser environments, where success rates fall to 14/25 and 12/25 at obstacle densities of 0.06 and 0.08, respectively. A likely reason is that YOPOv2 predicts a reference trajectory rather than direct low-level commands, and the desired acceleration must still be executed by a downstream SO(3) tracking controller. In dense, high-speed scenes, the available free-space margin shrinks rapidly, so small changes in primitive scoring can lead to abrupt switches between neighboring avoidance directions. In our rollouts, YOPOv2 sometimes changes from one side to the other when the quadrotor is already close to obstacles, leaving insufficient time for the low-level tracking controller to follow the updated reference smoothly, or gives maneuvering trajectory at extreme state. Such urgent planning changes near obstacles raise dynamic-infeasibility issues, resulting in coupled instable movement such as winding flight patterns. These issues degrade flight safety and ultimately cause elevated jerk and frequent failures in the hardest settings. Consequently, YOPOv2's primary limitation is not low-speed planning quality but rather predicting difficult trajectory for backend controller to track at high speeds.

Although the baselines described above differ in architecture, they share a common design pattern: each first generates an intermediate reference trajectory or command, which is then executed by a downstream low-level controller. This separation between high-level decision-making and low-level tracking performs well when flight speeds are moderate and sufficient correction margin is available. In dense, high-speed scenes, however, the control response becomes limited by the combined latency of perception, planning, and tracking. Consequently, the generated reference may already be outdated or difficult to track smoothly by the time it is executed. The resulting mismatch often leads to oscillatory corrections, elevated jerk, and eventual collisions. In contrast, Newton and our method adopt a differentiable end-to-end formulation that reduces the gap between decision and control, thereby improving consistency during aggressive flight.

Among the baselines, Newton achieves the highest overall success rate in the large-scale benchmark, with perfect performance in most sparse and moderate settings and 19 out of 25 successes even in the hardest cases. This strong performance is primarily due to its planning‑free design, which avoids the latency and error accumulation introduced by explicit trajectory generation and replanning. However, its remaining failures during aggressive flight in dense scenes appear to stem from two limitations.
First, Newton's policy predicts desired acceleration under a simplified point‑mass kinematics model, and the predicted command is then executed by an outer‑loop controller with yaw aligned to the target direction. While this abstraction simplifies training, it also weakens the perception–control coupling required for agile obstacle avoidance. In dense scenes, successful avoidance often demands coordinated changes in attitude and position, meanwhile heading to maintain both visibility and clearance. In our rollouts, Newton sometimes exhibits overly rapid pitch-up corrections for braking near obstacles, which abruptly twist the depth camera frame and destabilize the perception input. Once this instability persists for multiple steps, the resulting control response can become increasingly unreliable, eventually causing the vehicle to lose attitude stability and collide.
Second, Newton is trained under simplified point‑mass kinematics but evaluated with high‑fidelity dynamics, introducing a training–inference gap. This gap becomes more severe as flight velocity increases, because the backend controller is then more challenged to execute the predicted acceleration commands accurately—similar to the execution difficulty observed in YOPOv2. In dense scenes, such mismatch is particularly harmful during rapid consecutive avoidance maneuvers. If the resulting unstable motion lasts for several consecutive steps, recurrent inference can become increasingly unreliable, further increasing the risk of failure.
Newton thus remains highly robust overall, but its degradation in the hardest settings suggests that removing planning alone is insufficient without strong consistency between the learned commands and the execution during evaluation.

Overall, our method achieves the best performance across the large-scale benchmark when success rate, completion, forward progress, and smoothness are considered jointly. Unlike prior methods that first generate an intermediate reference and then rely on a downstream tracker, our policy directly maps perception to low-level CTBR commands. This design produces smooth actions, enabling natural auto‑acceleration and auto‑deceleration during flight while avoiding the abrupt corrective behavior often observed in dense, high‑speed scenes. Furthermore, the CTBR control interface provides a more direct perception‑aware coupling between observation and agile maneuvering, allowing the quadrotor to fly agilely while maintaining stable and clear visual feedback. These advantages are reflected in the results: our method maintains consistently high success and completion rates across all settings while achieving the lowest overall jerk, which remains within 2.97–23.19m/s³ even in the most aggressive regimes.

\subsubsection{Comparison in Super-dense Map}
To further evaluate robustness under denser clutter, we increase the obstacle density to 0.16 and 0.25 obstacles/m², where the free space between obstacles becomes extremely limited. Because these scenarios are substantially more constrained, we evaluate the policies at commanded velocities of 2, 4, 6, and 8m/s. The results are reported in Table~\ref{tab:comparison_dense}. The central question in this benchmark is whether a method can retain both reliability and non‑trivial forward progress once the free‑space margin becomes very small.

\begin{figure}
    \centering
    \includegraphics[width=1.0\linewidth]{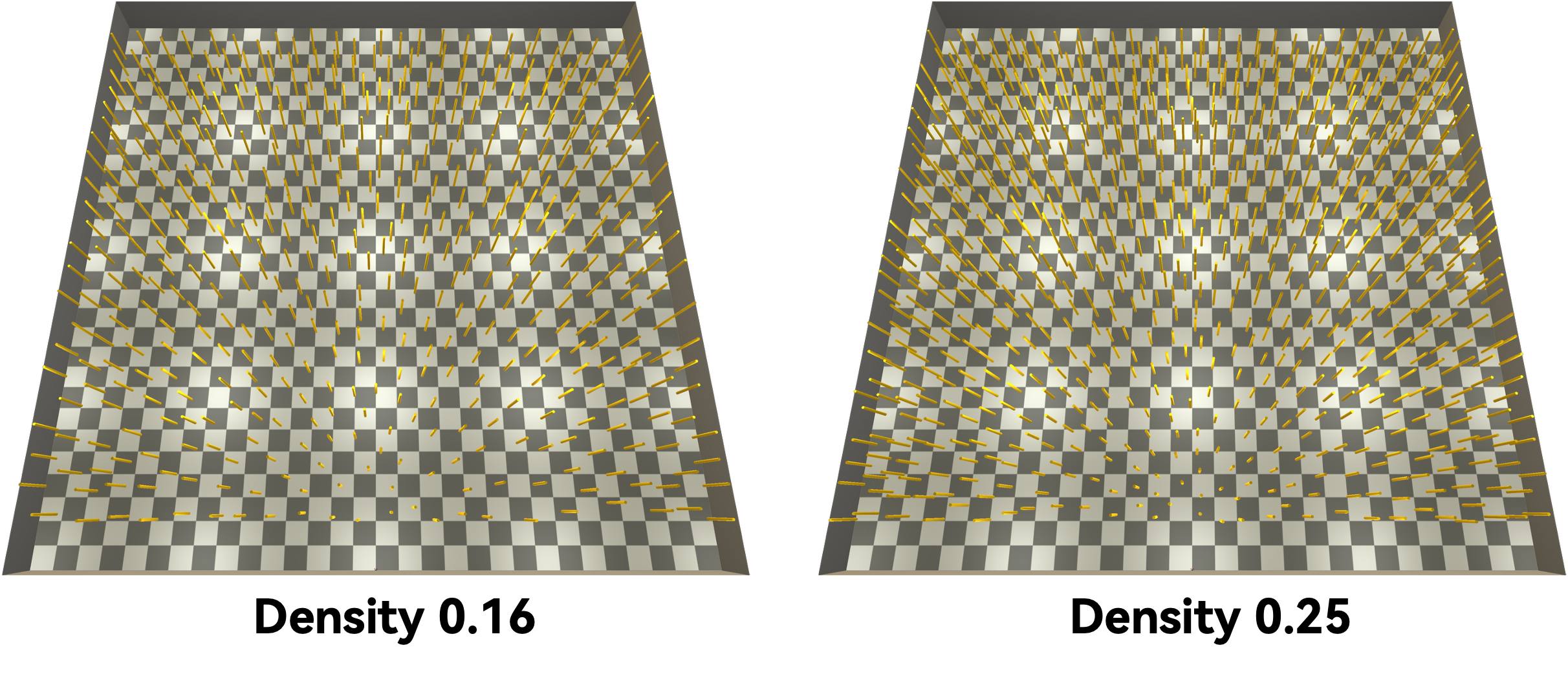}
    \caption{
        Panoramic views of the super-dense environments at two obstacle
        densities: 0.16 and 0.25 obstacles/m$^2$.
    }
    \label{fig:super_dense_scenes}
\end{figure}

\begin{table*}[ht!]
\centering
\setlength{\tabcolsep}{2.5pt} 
\small 
\renewcommand{\arraystretch}{1.2}
\caption{Key Character Comparison of Baselines in Superdense Scenes (Density 0.16 and 0.25) \label{tab:comparison_dense}}
\begin{tabular}{l | cccc | cccc | cccc | cccc | cccc}
\toprule

\multirow{2}{*}{\textbf{Velocity}} & 
\multicolumn{4}{c}{\textbf{EGO-Planner}} & 
\multicolumn{4}{c}{\textbf{Agile}} & 
\multicolumn{4}{c}{\textbf{Newton}} & 
\multicolumn{4}{c}{\textbf{YOPOv2}} & 
\multicolumn{4}{c}{\textbf{Ours}} \\
\cmidrule(lr){2-5} \cmidrule(lr){6-9} \cmidrule(lr){10-13} \cmidrule(lr){14-17} \cmidrule(lr){18-21}

& 
\textbf{S/T} & \textbf{Comp} & \textbf{Vel} & \textbf{Jerk} & 
\textbf{S/T} & \textbf{Comp} & \textbf{Vel} & \textbf{Jerk} & 
\textbf{S/T} & \textbf{Comp} & \textbf{Vel} & \textbf{Jerk} & 
\textbf{S/T} & \textbf{Comp} & \textbf{Vel} & \textbf{Jerk} &
\textbf{S/T} & \textbf{Comp} & \textbf{Vel} & \textbf{Jerk} \\
\midrule

\multicolumn{21}{c}{\textbf{Density: 0.16 obstacles/m$^2$}} \\
\midrule

{\textbf{2 m/s}} 
& \textbf{5/5} & \textbf{1.000} & 1.66 & 10.99 & 0/5 & 0.350 & 1.97 & 24.78 & \textbf{5/5} & \textbf{1.000} & 2.00 & 7.68 & 3/5 & 0.827 & 1.79 & \textbf{4.03} & \textbf{5/5} & \textbf{1.000} & \textbf{2.39} & 5.16 \\

{\textbf{4 m/s}} 
& 4/5 & 0.917 & 3.29 & 58.08 & 0/5 & 0.221 & 3.36 & 55.49 & \textbf{5/5} & \textbf{1.000} & 3.46 & 14.07 & 2/5 & 0.756 & 3.31 & 11.45 & \textbf{5/5} & \textbf{1.000} & \textbf{3.54} & \textbf{9.69} \\

{\textbf{6 m/s}} 
& 1/5 & 0.774 & 4.60 & 45.10 & 0/5 & 0.438 & \textbf{4.80} & 90.23 & \textbf{5/5} & \textbf{1.000} & 4.15 & 28.90 & 2/5 & 0.609 & 4.48 & 31.54 & \textbf{5/5} & \textbf{1.000} & 4.64 & \textbf{15.16} \\

{\textbf{8 m/s}} 
& 1/5 & 0.452 & 5.30 & 57.37 & 0/5 & 0.249 & 4.35 & 147.81 & 4/5 & 0.877 & 4.31 & 61.58 & 2/5 & 0.543 & \textbf{5.89} & 46.61 & \textbf{5/5} & \textbf{1.000} & 5.26 & \textbf{20.06} \\

\midrule

\multicolumn{21}{c}{\textbf{Density: 0.25 obstacles/m$^2$}} \\
\midrule

{\textbf{2 m/s}} 
& \textbf{5/5} & \textbf{1.000} & 1.65 & 16.05 & 0/5 & 0.259 & 1.89 & 36.08 & \textbf{5/5} & \textbf{1.000} & 1.75 & 8.66 & 2/5 & 0.728 & 1.75 & \textbf{3.57} & \textbf{5/5} & \textbf{1.000} & \textbf{2.24} & 6.25 \\

{\textbf{4 m/s}} 
& 2/5 & 0.664 & 3.14 & 103.57 & 0/5 & 0.219 & \textbf{3.26} & 61.21 & \textbf{5/5} & \textbf{1.000} & 2.68 & 19.51 & 1/5 & 0.485 & 3.03 & 12.70 & \textbf{5/5} & \textbf{1.000} & 3.11 & \textbf{10.31} \\

{\textbf{6 m/s}} 
& 0/5 & 0.305 & 3.66 & 91.34 & 0/5 & 0.231 & 3.95 & 100.71 & \textbf{5/5} & \textbf{1.000} & 3.23 & 30.70 & 0/5 & 0.487 & \textbf{4.63} & 21.45 & \textbf{5/5} & \textbf{1.000} & 3.87 & \textbf{16.92} \\

{\textbf{8 m/s}} 
& 0/5 & 0.267 & 4.60 & 134.36 & 0/5 & 0.164 & 3.55 & 139.21 & \textbf{5/5} & \textbf{1.000} & 3.15 & 35.93 & 0/5 & 0.556 & \textbf{6.03} & 68.55 & \textbf{5/5} & \textbf{1.000} & 4.39 & \textbf{21.90} \\

\bottomrule
\end{tabular}

\begin{tablenotes}
\item \textbf{S/T}: Success / Total Runs; \textbf{Comp}: Average Completion Rate; \textbf{Vel}: Average Velocity (m/s); \textbf{Jerk}: Average Jerk (m/s$^3$).
\item All velocity metrics, including average velocity and maximum velocity, are measured as the effective forward velocity along the goal direction, so detours around obstacles reduce the reported speed instead of inflating it.
\end{tablenotes}
\end{table*}

The super-dense benchmark reveals a substantially larger performance gap between methods. Agile fails in all tested configurations, indicating the limited robustness of its expert‑distilled trajectory policy in extremely cluttered spaces. EGO‑Planner also degrades rapidly as density increases; at a density of 0.25, it records zero successful runs once the commanded velocity reaches 6m/s. YOPOv2 retains partial completion in several cases, but its success rate becomes highly unstable and its jerk rises sharply, reaching 83.6m/s³ at a density of 0.25 and a velocity of 8m/s.

Newton remains highly robust in these super‑dense scenes, achieving 5/5 successes in every evaluated configuration, which confirms the strength of differentiable‑simulation training in this regime. However, this robustness comes at the cost of reduced forward velocity. Because the free space is extremely limited in these super‑dense environments, Newton often performs repeated braking for avoidance maneuvers in front of closely spaced obstacles, which severely limits its continuous forward progress. For example, at a density of 0.25 and a commanded velocity of 8m/s, its average forward speed is only 3.15m/s, indicating increasingly conservative motion as the free‑space margin narrows.

Our method also achieves 5/5 successes across all super‑dense settings while sustaining consistently higher forward velocity and lower jerk than Newton. At a density of 0.25 and 8m/s, our policy attains an average forward speed of 4.39m/s compared to Newton's 3.15m/s, while reducing jerk from 35.93 to 21.90m/s³. Similar gains appear in all other super‑dense configurations. These results indicate that, in this benchmark, our method retains a larger margin between robustness and useful forward progress than Newton.

\subsection{Real World Experiments}

\subsubsection{Hardware and Configuration}
As illustrated in \cref{fig:overview}, we employ a custom-built quadrotor platform for real-world experiments. The platform is equipped with a high-performance onboard computer (Nvidia Jetson Orin NX), a stereo camera (Intel RealSense D435i), and a Betaflight flight controller. The drone is designed to be lightweight and agile, enabling high-speed navigation through complex environments. We further refine the component layout to align the center of gravity with the geometric center, thereby improving control performance. For safety considerations, the maximum thrust is limited to twice the drone's weight.
During flight, all computational tasks—including state estimation, control, and policy inference—are executed onboard to guarantee real-time responsiveness, without relying on any external computation or communication. The quadrotor obtains its state information from a visual-inertial odometry (VIO) module following \cite{qin2018vins}.


\subsubsection{Outdoor Experiments}
We conduct extensive outdoor experiments to evaluate the performance of our proposed policy in real-world scenarios. The experiments are designed to assess the drone's ability to navigate diverse environments, including urban settings, natural landscapes, and complex obstacle courses.

\begin{figure*}
    \centering
    \includegraphics[width=1.0\textwidth]{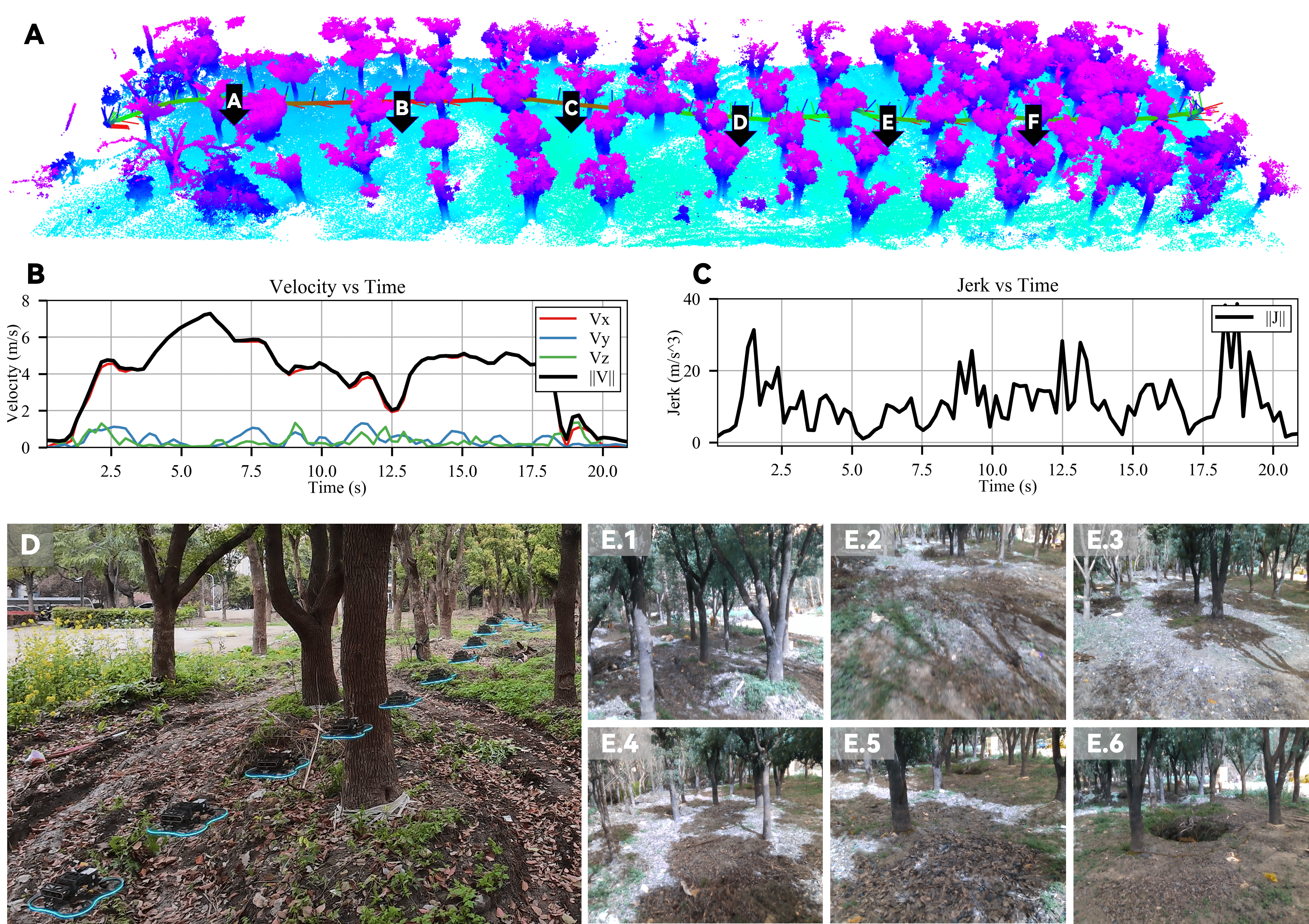}
    \caption{
        Real-world experiment in a regular forest environment with an obstacle density of 0.1~m$^{-2}$. \textbf{(A)} Reconstructed scene and flight trajectory. \textbf{(B)} Velocity profile during flight. \textbf{(C)} Jerk profile during flight. \textbf{(D)} Real-world scene and flight trajectory. \textbf{(E)} First-person view of the drone at the time steps indicated in (A).
    }
    \label{fig:fast}
\end{figure*}

We first evaluate the policy in a regular forest with an obstacle density of $0.1~\text{m}^{-2}$, where the drone must navigate through trees. Given a target located $90~\text{m}$ ahead of the starting point and a maximum commanded velocity of $10~\text{m/s}$, the quadrotor is required to traverse the forest safely under the trained policy. The reconstructed flight trajectory and the corresponding scene are presented in \cref{fig:fast}.A. As shown in \cref{fig:fast}.B and C, the drone successfully maneuvers through the trees at high speeds, demonstrating its ability to handle complex environments. The maximum speed attained in this test is approximately $7.5~\text{m/s}$—comparable to state-of-the-art methods, despite the simplicity of our architecture.
Unlike approaches that rely on precomputed trajectories and force a low-level controller to track them, our policy exhibits strong velocity adaptation in response to local obstacle density. This enables the drone to maintain safety while navigating at high speeds.

\begin{figure*}
    \centering
    \includegraphics[width=1.0\textwidth]{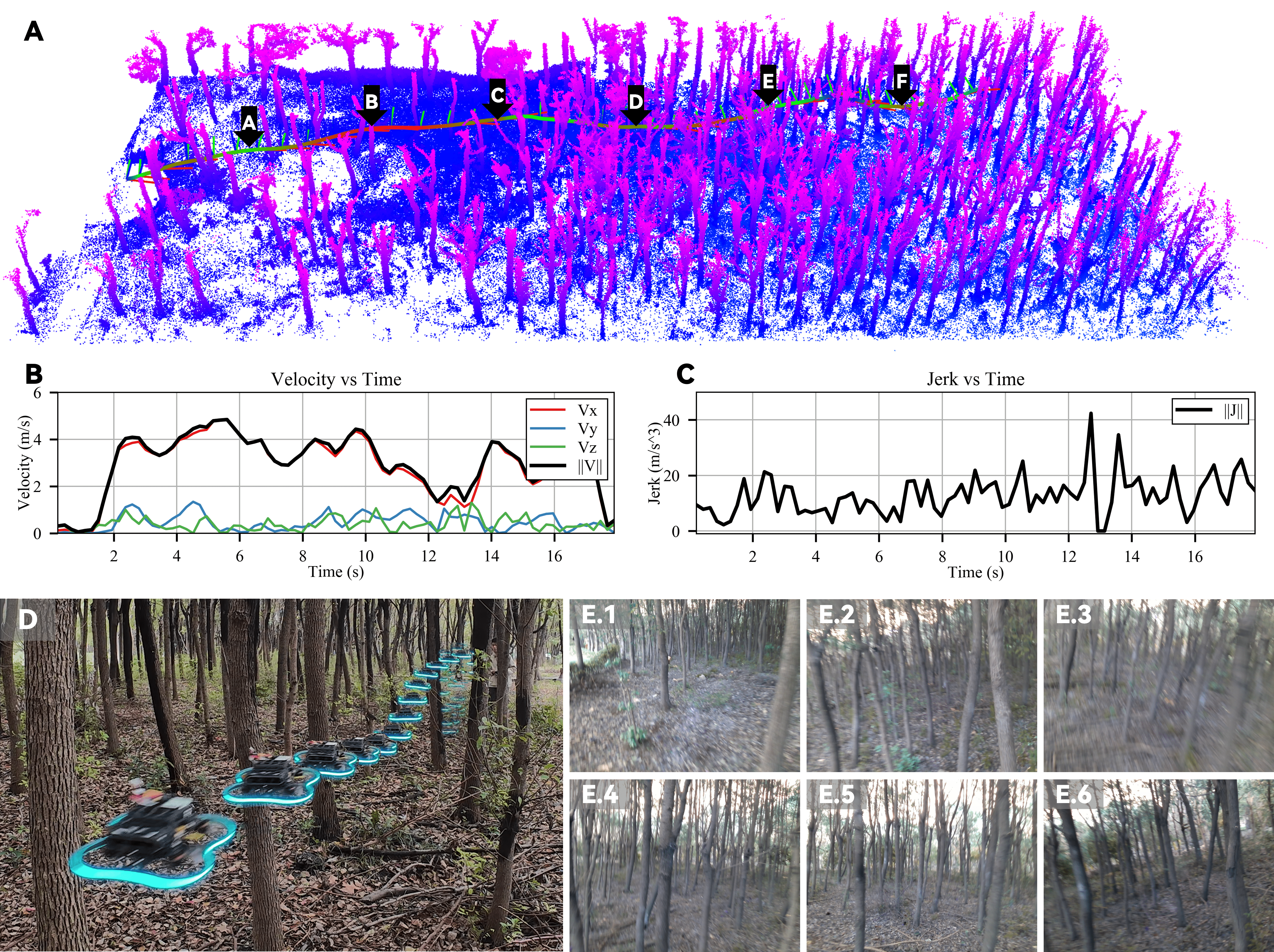}
    \caption{        
        Real-world experiment in a super-dense forest with an obstacle density of approximately $1~\text{m}^{-2}$. \textbf{(A)} Reconstructed three-dimensional scene and the actual flight trajectory. \textbf{(B)} Velocity profile recorded during the flight. \textbf{(C)} Jerk profile recorded during the flight. \textbf{(D)} Photograph of the real-world environment alongside the corresponding flight trajectory. \textbf{(E)} First-person view (FPV) captured at the time steps indicated in \textbf{(A)}.
    }
    \label{fig:dense}
\end{figure*}

To further assess the advantage of our end-to-end architecture, we evaluate the policy in a super-dense forest with an approximate obstacle density of $1~\text{m}^{-2}$, as depicted in \cref{fig:dense}.A and D. To the best of our knowledge, no existing work has been tested in such a challenging scenario. In relatively open areas, the quadrotor reaches a peak speed of approximately $5~\text{m/s}$, whereas in the densest regions it autonomously reduces its average speed to around $3~\text{m/s}$. Moreover, the policy exhibits a strong ability to maintain perception awareness while weaving through dense trees—a critical capability for collision-free navigation in such environments. As shown in \cref{fig:dense}.E, the field of view (FOV) consistently remains oriented forward. In contrast, the real-world deployment of Newton exhibits severe vertical oscillations, which cause its FOV to periodically point toward the ground, lose sight of the target, and ultimately result in a crash.

\begin{figure*}
    \centering
    \includegraphics[width=1.0\textwidth]{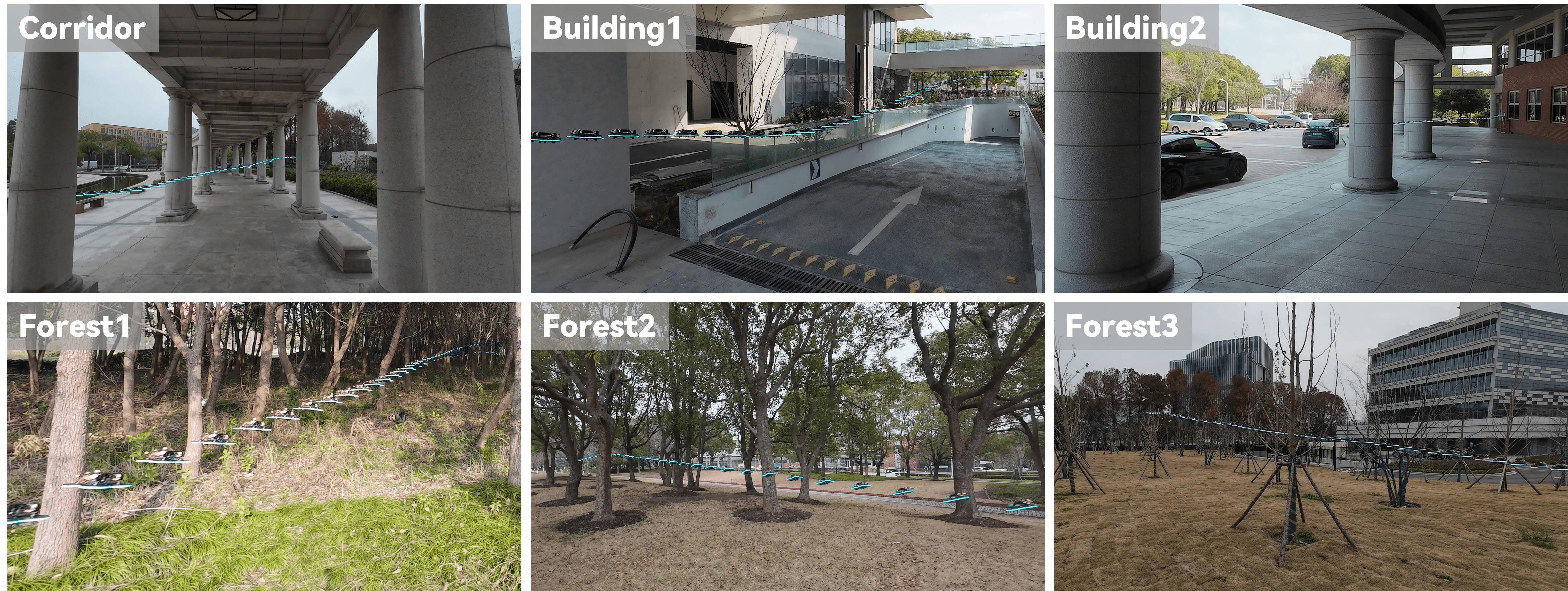}
    \caption{Multiple experiments for testing policy generalization in various environments including wild and urban scenes.}
    \label{fig:robust}
\end{figure*}

Moreover, to validate the generalization of the proposed policy, we conduct multiple experiments in diverse environments, including both wild and urban scenes, as shown in \cref{fig:robust}. The results demonstrate that the policy adapts well to different scenarios and maintains robust performance, highlighting its potential for real-world deployment across a variety of settings.

Unlike previous methods that rely on trajectory tracking under a point-mass model, our low-level control architecture enables full authority over the drone's dynamics. This allows the drone to avoid collisions with only minimal adjustments, rather than requiring large, aggressive maneuvers—resulting in a safer and more efficient behavior. Such elegant, minimal intervention also benefits perception-aware requirements. Consequently, as illustrated in \cref{fig:robust}, \cref{fig:dense}.A, and \cref{fig:fast}.A, the resulting flight trajectories are smooth and nearly straight. Although the drone may appear unresponsive to the environment, it in fact performs only the subtle adjustments necessary to avoid obstacles.

\section{Discussion}
Our work achieves state-of-the-art overall performance in balancing safety, stability, and agility, despite using the simplest training and inference pipeline. It inherits the advantages of Newton while addressing its key drawbacks: loss of perception due to urgent braking, instability caused by dynamics-infeasible commands, and the training-to-inference gap. By issuing low-level commands, our method fully leverages the drone's control capacity, transforming aggressive maneuvering into subtle adjustments during collision avoidance. However, it shares a common limitation with mapping-free collision-avoidance baselines: a lack of planning ability in complex environments such as mazes. Consequently, such policies are best suited as basic safety controllers, complemented by an external front-end planner that provides high-level directional commands.
This work is released at \url{https://github.com/Fanxing-LI/avoidance}.

\section{Conclusion}
In this paper, we train an end-to-end policy for collision-free flight using reinforcement learning via differentiable simulation. The policy directly maps depth images to low-level control commands, enabling full quadrotor control and avoiding dynamics-infeasible issues. Reinforcement learning with differentiable simulation provides highly precise analytical gradients for optimizing the actor, thereby overcoming the difficulty of training end-to-end policies.

We conduct extensive experiments to evaluate our method both in simulation and the real world. In simulation, our policy outperforms other methods in terms of safety, stability, and average velocity across various obstacle densities and scenes. In real-world experiments, despite the simplicity of its training and inference pipeline, our method achieves state-of-the-art performance, reaching a maximum velocity of 7.5~m/s in a regular forest. Furthermore, it stably attains a velocity of 5.0~m/s in a super-dense forest while maintaining perception awareness.

Our policy performs well without relying on any external modules—such as front-end or back-end controllers, mapping, recurrent architectures, complex backbones, or action primitives—nor on training tricks like curriculum learning or privileged guidance.
The simplicity of our pipeline, combined with the power of differentiable simulation, demonstrates that effective collision-free flight does not require increasingly complex modules or heuristics. Instead, directly optimizing a low-level control policy with analytical gradients offers a clean, efficient, and generalizable solution for agile aerial navigation.

\bibliographystyle{IEEEtran}
\bibliography{reference.bib}

@article{jung_perception_2018,
	title = {Perception, guidance, and navigation for indoor autonomous drone racing using deep learning},
	volume = {3},
	issn = {2377-3766},
	number = {3},
	journal = {IEEE Robotics and Automation Letters},
	author = {Jung, Sunggoo and Hwang, Sunyou and Shin, Heemin and Shim, David Hyunchul},
	year = {2018},
	pages = {2539--2544},
}

@inproceedings{kaufmann_beauty_2019,
	title = {Beauty and the beast: {Optimal} methods meet learning for drone racing},
	isbn = {1-5386-6027-X},
	publisher = {IEEE},
	author = {Kaufmann, Elia and Gehrig, Mathias and Foehn, Philipp and Ranftl, René and Dosovitskiy, Alexey and Koltun, Vladlen and Scaramuzza, Davide},
	year = {2019},
	pages = {690--696},
}

@article{cabrera-ponce_gate_2019,
	title = {Gate detection for micro aerial vehicles using a single shot detector},
	volume = {17},
	issn = {1548-0992},
	number = {12},
	journal = {IEEE Latin America Transactions},
	author = {Cabrera-Ponce, Aldrich A. and Rojas-Perez, Leticia Oyuki and Carrasco-Ochoa, Jesus Ariel and Martinez-Trinidad, Jose Francisco and Martinez-Carranza, Jose},
	year = {2019},
	pages = {2045--2052},
}

@article{hanover_autonomous_2024,
	title = {Autonomous {Drone} {Racing}: {A} {Survey}},
	volume = {40},
	issn = {1941-0468},
	shorttitle = {Autonomous {Drone} {Racing}},
	url = {https://ieeexplore.ieee.org/document/10530312},
	doi = {10.1109/TRO.2024.3400838},
	urldate = {2024-07-01},
	journal = {IEEE Transactions on Robotics},
	author = {Hanover, Drew and Loquercio, Antonio and Bauersfeld, Leonard and Romero, Angel and Penicka, Robert and Song, Yunlong and Cioffi, Giovanni and Kaufmann, Elia and Scaramuzza, Davide},
	year = {2024},
	note = {Conference Name: IEEE Transactions on Robotics},
	pages = {3044--3067},
}

@article{giusti_machine_2015,
	title = {A machine learning approach to visual perception of forest trails for mobile robots},
	volume = {1},
	issn = {2377-3766},
	number = {2},
	journal = {IEEE Robotics and Automation Letters},
	author = {Giusti, Alessandro and Guzzi, Jérôme and Cireşan, Dan C. and He, Fang-Lin and Rodríguez, Juan P. and Fontana, Flavio and Faessler, Matthias and Forster, Christian and Schmidhuber, Jürgen and Di Caro, Gianni},
	year = {2015},
	pages = {661--667},
}

@article{loquercio_dronet_2018,
	title = {Dronet: {Learning} to fly by driving},
	volume = {3},
	issn = {2377-3766},
	number = {2},
	journal = {IEEE Robotics and Automation Letters},
	author = {Loquercio, Antonio and Maqueda, Ana I. and Del-Blanco, Carlos R. and Scaramuzza, Davide},
	year = {2018},
	pages = {1088--1095},
}

@article{loquercio_deep_2019,
	title = {Deep drone racing: {From} simulation to reality with domain randomization},
	volume = {36},
	issn = {1552-3098},
	number = {1},
	journal = {IEEE Transactions on Robotics},
	author = {Loquercio, Antonio and Kaufmann, Elia and Ranftl, René and Dosovitskiy, Alexey and Koltun, Vladlen and Scaramuzza, Davide},
	year = {2019},
	pages = {1--14},
}

@inproceedings{kaufmann_deep_2018,
	title = {Deep drone racing: {Learning} agile flight in dynamic environments},
	isbn = {2640-3498},
	publisher = {PMLR},
	author = {Kaufmann, Elia and Loquercio, Antonio and Ranftl, Rene and Dosovitskiy, Alexey and Koltun, Vladlen and Scaramuzza, Davide},
	year = {2018},
	pages = {133--145},
}

@inproceedings{gandhi_learning_2017,
	title = {Learning to fly by crashing},
	isbn = {1-5386-2682-9},
	publisher = {IEEE},
	author = {Gandhi, Dhiraj and Pinto, Lerrel and Gupta, Abhinav},
	year = {2017},
	pages = {3948--3955},
}

@article{r_penicka_learning_2022,
	title = {Learning {Minimum}-{Time} {Flight} in {Cluttered} {Environments}},
	volume = {7},
	issn = {2377-3766},
	doi = {10.1109/LRA.2022.3181755},
	number = {3},
	journal = {IEEE Robotics and Automation Letters},
	author = {R. Penicka and Y. Song and E. Kaufmann and D. Scaramuzza},
	year = {2022},
	pages = {7209--7216},
}

@article{loquercio_learning_2021,
	title = {Learning high-speed flight in the wild},
	volume = {6},
	issn = {2470-9476},
	number = {59},
	journal = {Science Robotics},
	author = {Loquercio, Antonio and Kaufmann, Elia and Ranftl, René and Müller, Matthias and Koltun, Vladlen and Scaramuzza, Davide},
	year = {2021},
	pages = {eabg5810},
}

@inproceedings{kouris_learning_2018,
	title = {Learning to fly by myself: {A} self-supervised cnn-based approach for autonomous navigation},
	isbn = {1-5386-8094-7},
	publisher = {IEEE},
	author = {Kouris, Alexandros and Bouganis, Christos-Savvas},
	year = {2018},
	pages = {1--9},
}

@article{song_reaching_2023,
	title = {Reaching the limit in autonomous racing: {Optimal} control versus reinforcement learning},
	volume = {8},
	doi = {doi:10.1126/scirobotics.adg1462},
	number = {82},
	journal = {Science Robotics},
	author = {Song, Yunlong and Romero, Angel and Müller, Matthias and Koltun, Vladlen and Scaramuzza, Davide},
	year = {2023},
	pages = {eadg1462},
}

@article{yu_mavrl_2024,
	title = {{MAVRL}: Learn to Fly in Cluttered Environments with Varying Speed},
	journaltitle = {{arXiv} preprint {arXiv}:2402.08381},
	author = {Yu, Hang and De Wagter, Christophe and de Croon, Guido {CH}},
	date = {2024},
}

@misc{lu_yopov2-tracker_2025,
	title = {{YOPOv}2-Tracker: An End-to-End Agile Tracking and Navigation Framework from Perception to Action},
	url = {http://arxiv.org/abs/2505.06923},
	doi = {10.48550/arXiv.2505.06923},
	shorttitle = {{YOPOv}2-Tracker},
	number = {{arXiv}:2505.06923},
	publisher = {{arXiv}},
	author = {Lu, Junjie and Hui, Yulin and Zhang, Xuewei and Feng, Wencan and Shen, Hongming and Li, Zhiyu and Tian, Bailing},
	urldate = {2025-07-09},
	date = {2025-05-11},
	eprinttype = {arxiv},
	eprint = {2505.06923 [cs]},
}

@inproceedings{y_song_learning_2023,
	title = {Learning Perception-Aware Agile Flight in Cluttered Environments},
	doi = {10.1109/ICRA48891.2023.10160563},
	eventtitle = {2023 {IEEE} International Conference on Robotics and Automation ({ICRA})},
	pages = {1989--1995},
	author = {Y. Song and K. Shi and R. Penicka and D. Scaramuzza},
	date = {2023-06-29},
}

@misc{zhang_back_2024,
	title = {Back to Newton's Laws: Learning Vision-based Agile Flight via Differentiable Physics},
	url = {http://arxiv.org/abs/2407.10648},
	doi = {10.48550/arXiv.2407.10648},
	shorttitle = {Back to Newton's Laws},
	number = {{arXiv}:2407.10648},
	publisher = {{arXiv}},
	author = {Zhang, Yuang and Hu, Yu and Song, Yunlong and Zou, Danping and Lin, Weiyao},
	urldate = {2025-05-01},
	date = {2024-07-16},
	eprinttype = {arxiv},
	eprint = {2407.10648 [cs]},
}

@misc{li_abpt_2025,
	title = {{ABPT}: {Amended} {Backpropagation} through {Time} with {Partially} {Differentiable} {Rewards}},
	shorttitle = {{ABPT}},
	url = {http://arxiv.org/abs/2501.14513},
	doi = {10.48550/arXiv.2501.14513},
	urldate = {2025-08-22},
	publisher = {arXiv},
	author = {Li, Fanxing and Sun, Fangyu and Zhang, Tianbao and Zou, Danping},
	month = may,
	year = {2025},
	note = {arXiv:2501.14513 [cs]},
}

@misc{heeg_learning_2024,
	title = {Learning {Quadrotor} {Control} {From} {Visual} {Features} {Using} {Differentiable} {Simulation}},
	url = {http://arxiv.org/abs/2410.15979},
	doi = {10.48550/arXiv.2410.15979},
	urldate = {2024-10-22},
	publisher = {arXiv},
	author = {Heeg, Johannes and Song, Yunlong and Scaramuzza, Davide},
	month = oct,
	year = {2024},
	note = {arXiv:2410.15979},
}

@misc{kim_rapid_2025,
	title = {{RAPID}: Robust and Agile Planner Using Inverse Reinforcement Learning for Vision-Based Drone Navigation},
	url = {http://arxiv.org/abs/2502.02054},
	doi = {10.48550/arXiv.2502.02054},
	shorttitle = {{RAPID}},
	number = {{arXiv}:2502.02054},
	publisher = {{arXiv}},
	author = {Kim, Minwoo and Bae, Geunsik and Lee, Jinwoo and Shin, Woojae and Kim, Changseung and Choi, Myong-Yol and Shin, Heejung and Oh, Hyondong},
	urldate = {2025-02-08},
	date = {2025-02-04},
	eprinttype = {arxiv},
	eprint = {2502.02054 [cs]},
}

@inproceedings{bhattacharya_vision_2025,
	title = {Vision Transformers for End-to-End Vision-Based Quadrotor Obstacle Avoidance},
	url = {https://ieeexplore.ieee.org/abstract/document/11128042},
	doi = {10.1109/ICRA55743.2025.11128042},
	eventtitle = {2025 {IEEE} International Conference on Robotics and Automation ({ICRA})},
	pages = {1--8},
	booktitle = {2025 {IEEE} International Conference on Robotics and Automation ({ICRA})},
	author = {Bhattacharya, Anish and Rao, Nishanth and Parikh, Dhruv and Kunapuli, Pratik and Wu, Yuwei and Tao, Yuezhan and Matni, Nikolai and Kumar, Vijay},
	urldate = {2026-01-20},
	date = {2025-05},
}

@inproceedings{kulkarni_reinforcement_2024,
	title = {Reinforcement Learning for Collision-free Flight Exploiting Deep Collision Encoding},
	url = {https://ieeexplore.ieee.org/abstract/document/10610287},
	doi = {10.1109/ICRA57147.2024.10610287},
	eventtitle = {2024 {IEEE} International Conference on Robotics and Automation ({ICRA})},
	pages = {15781--15788},
	booktitle = {2024 {IEEE} International Conference on Robotics and Automation ({ICRA})},
	author = {Kulkarni, Mihir and Alexis, Kostas},
	urldate = {2026-01-20},
	date = {2024-05},
}

@article{b_zhou_robust_2019,
	title = {Robust and Efficient Quadrotor Trajectory Generation for Fast Autonomous Flight},
	volume = {4},
	issn = {2377-3766},
	doi = {10.1109/LRA.2019.2927938},
	pages = {3529--3536},
	number = {4},
	journaltitle = {{IEEE} Robotics and Automation Letters},
	author = {B. Zhou and F. Gao and L. Wang and C. Liu and S. Shen},
	date = {2019},
}

@inproceedings{zhou_ego-swarm_2021,
	title = {{EGO}-Swarm: A Fully Autonomous and Decentralized Quadrotor Swarm System in Cluttered Environments},
	url = {https://ieeexplore.ieee.org/abstract/document/9561902},
	doi = {10.1109/ICRA48506.2021.9561902},
	shorttitle = {{EGO}-Swarm},
	eventtitle = {2021 {IEEE} International Conference on Robotics and Automation ({ICRA})},
	pages = {4101--4107},
	booktitle = {2021 {IEEE} International Conference on Robotics and Automation ({ICRA})},
	author = {Zhou, Xin and Zhu, Jiangchao and Zhou, Hongyu and Xu, Chao and Gao, Fei},
	urldate = {2026-01-28},
	date = {2021-05},
	note = {{ISSN}: 2577-087X},
}

@inproceedings{richter_polynomial_2016,
	title = {Polynomial trajectory planning for aggressive quadrotor flight in dense indoor environments},
	isbn = {3-319-28870-9},
	eventtitle = {Robotics Research: The 16th International Symposium {ISRR}},
	pages = {649--666},
	publisher = {Springer},
	author = {Richter, Charles and Bry, Adam and Roy, Nicholas},
	date = {2016},
}

@inproceedings{allen_real-time_2016,
	title = {A real-time framework for kinodynamic planning with application to quadrotor obstacle avoidance},
	eventtitle = {{AIAA} Guidance, Navigation, and Control Conference},
	pages = {1374},
	author = {Allen, Ross and Pavone, Marco},
	date = {2016},
}

@article{s_liu_search-based_2018,
	title = {Search-Based Motion Planning for Aggressive Flight in {SE}(3)},
	volume = {3},
	issn = {2377-3766},
	doi = {10.1109/LRA.2018.2795654},
	pages = {2439--2446},
	number = {3},
	journaltitle = {{IEEE} Robotics and Automation Letters},
	author = {S. Liu and K. Mohta and N. Atanasov and V. Kumar},
	date = {2018},
}

@article{zhou_ego-planner_2021,
	title = {{EGO}-Planner: An {ESDF}-Free Gradient-Based Local Planner for Quadrotors},
	volume = {6},
	issn = {2377-3766},
	url = {https://ieeexplore.ieee.org/abstract/document/9309347},
	doi = {10.1109/LRA.2020.3047728},
	shorttitle = {{EGO}-Planner},
	pages = {478--485},
	number = {2},
	journaltitle = {{IEEE} Robotics and Automation Letters},
	author = {Zhou, Xin and Wang, Zhepei and Ye, Hongkai and Xu, Chao and Gao, Fei},
	urldate = {2026-01-22},
	date = {2021-04},
}

@article{han_hierarchically_2025,
	title = {Hierarchically depicting vehicle trajectory with stability in complex environments},
	rights = {Copyright © 2025 The Authors, some rights reserved; exclusive licensee American Association for the Advancement of Science. No claim to original U.S. Government Works},
	url = {https://www.science.org/doi/10.1126/scirobotics.ads4551},
	doi = {10.1126/scirobotics.ads4551},
	journaltitle = {Science Robotics},
	author = {Han, Zhichao and Tian, Mengze and Gongye, Zaitian and Xue, Donglai and Xing, Jiaxi and Wang, Qianhao and Gao, Yuman and Wang, Jingping and Xu, Chao and Gao, Fei},
	urldate = {2026-01-28},
	date = {2025-06-18},
	note = {Publisher: American Association for the Advancement of Science},
}

@article{kaufmann_champion-level_2023,
	title = {Champion-level drone racing using deep reinforcement learning},
	volume = {620},
	issn = {1476-4687},
	doi = {10.1038/s41586-023-06419-4},
	pages = {982--987},
	number = {7976},
	journaltitle = {Nature},
	author = {Kaufmann, Elia and Bauersfeld, Leonard and Loquercio, Antonio and Müller, Matthias and Koltun, Vladlen and Scaramuzza, Davide},
	date = {2023-08-01},
}

@incollection{sutton1990integrated,
  title={Integrated architectures for learning, planning, and reacting based on approximating dynamic programming},
  author={Sutton, Richard S},
  booktitle={Machine learning proceedings 1990},
  pages={216--224},
  year={1990},
  publisher={Elsevier}
}

@article{janner2019trust,
  title={When to trust your model: Model-based policy optimization},
  author={Janner, Michael and Fu, Justin and Zhang, Marvin and Levine, Sergey},
  journal={Advances in neural information processing systems},
  volume={32},
  year={2019}
}

@article{chua2018deep,
  title={Deep reinforcement learning in a handful of trials using probabilistic dynamics models},
  author={Chua, Kurtland and Calandra, Roberto and McAllister, Rowan and Levine, Sergey},
  journal={Advances in neural information processing systems},
  volume={31},
  year={2018}
}

@inproceedings{nagabandi2018neural,
  title={Neural network dynamics for model-based deep reinforcement learning with model-free fine-tuning},
  author={Nagabandi, Anusha and Kahn, Gregory and Fearing, Ronald S and Levine, Sergey},
  booktitle={2018 IEEE international conference on robotics and automation (ICRA)},
  pages={7559--7566},
  year={2018},
  organization={IEEE}
}

@article{hafner2019dream,
  title={Dream to control: Learning behaviors by latent imagination},
  author={Hafner, Danijar and Lillicrap, Timothy and Ba, Jimmy and Norouzi, Mohammad},
  journal={arXiv preprint arXiv:1912.01603},
  year={2019}
}

@article{schulman2017proximal,
  title={Proximal policy optimization algorithms},
  author={Schulman, John and Wolski, Filip and Dhariwal, Prafulla and Radford, Alec and Klimov, Oleg},
  journal={arXiv preprint arXiv:1707.06347},
  year={2017}
}

@article{haarnoja2018soft,
  title={Soft actor-critic algorithms and applications},
  author={Haarnoja, Tuomas and Zhou, Aurick and Hartikainen, Kristian and Tucker, George and Ha, Sehoon and Tan, Jie and Kumar, Vikash and Zhu, Henry and Gupta, Abhishek and Abbeel, Pieter and others},
  journal={arXiv preprint arXiv:1812.05905},
  year={2018}
}

@inproceedings{mora_pods_2021,
	title = {{PODS}: {Policy} {Optimization} via {Differentiable} {Simulation}},
	shorttitle = {{PODS}},
	url = {https://proceedings.mlr.press/v139/mora21a.html},
	language = {en},
	urldate = {2024-11-17},
	booktitle = {Proceedings of the 38th {International} {Conference} on {Machine} {Learning}},
	publisher = {PMLR},
	author = {Mora, Miguel Angel Zamora and Peychev, Momchil and Ha, Sehoon and Vechev, Martin and Coros, Stelian},
	month = jul,
	year = {2021},
	note = {ISSN: 2640-3498},
	pages = {7805--7817},
}

@inproceedings{suh_differentiable_2022,
	title = {Do {Differentiable} {Simulators} {Give} {Better} {Policy} {Gradients}?},
	url = {https://proceedings.mlr.press/v162/suh22b.html},
	language = {en},
	urldate = {2024-11-17},
	booktitle = {Proceedings of the 39th {International} {Conference} on {Machine} {Learning}},
	publisher = {PMLR},
	author = {Suh, Hyung Ju and Simchowitz, Max and Zhang, Kaiqing and Tedrake, Russ},
	month = jun,
	year = {2022},
	note = {ISSN: 2640-3498},
	pages = {20668--20696},
}

@inproceedings{zhang_adaptive_2023,
	title = {Adaptive {Barrier} {Smoothing} for {First}-{Order} {Policy} {Gradient} with {Contact} {Dynamics}},
	url = {https://proceedings.mlr.press/v202/zhang23s.html},
	language = {en},
	urldate = {2024-11-22},
	booktitle = {Proceedings of the 40th {International} {Conference} on {Machine} {Learning}},
	publisher = {PMLR},
	author = {Zhang, Shenao and Jin, Wanxin and Wang, Zhaoran},
	month = jul,
	year = {2023},
	note = {ISSN: 2640-3498},
	pages = {41219--41243},
}

@misc{li_visfly_2024,
	title = {{VisFly}: {An} {Efficient} and {Versatile} {Simulator} for {Training} {Vision}-based {Flight}},
	shorttitle = {{VisFly}},
	url = {http://arxiv.org/abs/2407.14783},
	doi = {10.48550/arXiv.2407.14783},

	urldate = {2024-11-27},
	publisher = {arXiv},
	author = {Li, Fanxing and Sun, Fangyu and Zhang, Tianbao and Zou, Danping},
	month = sep,
	year = {2024},
	note = {arXiv:2407.14783},
}

@article{li2025abpt,
  title={ABPT: Amended Backpropagation through Time with Partially Differentiable Rewards},
  author={Li, Fanxing and Sun, Fangyu and Zhang, Tianbao and Zou, Danping},
  journal={arXiv preprint arXiv:2501.14513},
  year={2025}
}

@article{pan2026learning,
  title={Learning on the Fly: Rapid Policy Adaptation via Differentiable Simulation},
  author={Pan, Jiahe and Xing, Jiaxu and Reiter, Rudolf and Zhai, Yifan and Aljalbout, Elie and Scaramuzza, Davide},
  journal={IEEE Robotics and Automation Letters},
  year={2026},
  publisher={IEEE}
}

@misc{xu_accelerated_2022,
	title = {Accelerated {Policy} {Learning} with {Parallel} {Differentiable} {Simulation}},
	url = {http://arxiv.org/abs/2204.07137},
	doi = {10.48550/arXiv.2204.07137},
	urldate = {2024-10-12},
	publisher = {arXiv},
	author = {Xu, Jie and Makoviychuk, Viktor and Narang, Yashraj and Ramos, Fabio and Matusik, Wojciech and Garg, Animesh and Macklin, Miles},
	month = apr,
	year = {2022},
	note = {arXiv:2204.07137},
}

@article{raffin2021stable,
  title={Stable-baselines3: Reliable reinforcement learning implementations},
  author={Raffin, Antonin and Hill, Ashley and Gleave, Adam and Kanervisto, Anssi and Ernestus, Maximilian and Dormann, Noah},
  journal={Journal of machine learning research},
  volume={22},
  number={268},
  pages={1--8},
  year={2021}
}

@article{qin2018vins,
  title={Vins-mono: A robust and versatile monocular visual-inertial state estimator},
  author={Qin, Tong and Li, Peiliang and Shen, Shaojie},
  journal={IEEE transactions on robotics},
  volume={34},
  number={4},
  pages={1004--1020},
  year={2018},
  publisher={IEEE}
}


 




\vfill

\end{document}